\documentclass[11pt]{article}

\usepackage[final]{acl}

\usepackage{times}
\usepackage{latexsym}

\usepackage[T1]{fontenc}

\usepackage[utf8]{inputenc}

\usepackage{microtype}

\usepackage{inconsolata}

\usepackage{graphicx}

\usepackage{kotex}   
\usepackage{comment}
\usepackage{amssymb}

\usepackage{booktabs}

\usepackage{tcolorbox}
\tcbuselibrary{breakable}
\tcbset{
  mybox/.style={
    colback=gray!5,
    colframe=black!60,
    boxrule=0.5pt,
    arc=2mm,
    outer arc=2mm,
    left=6pt,
    right=6pt,
    top=6pt,
    bottom=6pt,
    breakable
  }
}

\usepackage{graphicx}
\usepackage{wrapfig}

\usepackage{enumitem}
\usepackage[utf8]{inputenc}

\usepackage{booktabs}
\usepackage{xcolor}

\usepackage{subcaption}
\newcommand{\ours}{\textsc{Mentor}}

\usepackage{array}
\usepackage{makecell}
\usepackage{geometry}
\usepackage{amsmath}
\usepackage{booktabs, tabularx, makecell, threeparttable}
\usepackage{amsmath}

\usepackage[dvipsnames]{xcolor}  

\usepackage{multirow}

\usepackage{algorithm}
\usepackage{algpseudocode}

\usepackage[table]{xcolor}
\usepackage{adjustbox}

\usepackage{booktabs}
\usepackage[table]{xcolor}
\usepackage{tabularx}
\usepackage{ragged2e}
\usepackage[table]{xcolor} 
\usepackage{colortbl}     

\tcbuselibrary{listings} 

\usepackage{listings}

\title{\ours{}: Reinforcement Learning via Flexible Teacher-Optimized Rewards for Tool-Use Distillation}

\newcommand\CoauthorMark{\footnotemark[\arabic{footnote}]}
\author{ChangSu Choi$^1$\thanks{Equal Contribution}\hspace{1.5mm} Hoyun Song$^{2,1}$\CoauthorMark\hspace{1.5mm} Dongyeon Kim$^1$\hspace{1.5mm} Minkyung Cho$^1$\hspace{1.5mm} WooHyeon Jung$^1$\\
\bf Sunjin Park$^3$\hspace{1.5mm} NohHyeob Bae$^3$\hspace{1.5mm} Seona Yu$^3$\hspace{1.5mm} KyungTae Lim$^1$\thanks{Corresponding Author} \\
$^1$Korea Advanced Institute of Science and Technology (KAIST)\\
$^2$Department of Artificial Intelligence, Dankook University\hspace{1.5mm} $^3$LG CNS\\
\hspace{1.5mm} \texttt{\{choics2623, ktlim\}@kaist.ac.kr} \hspace{1.5mm} \texttt{hoyun.song@dankook.ac.kr}
}

\begin{document}
\maketitle

\begin{abstract}
Distilling the tool-use capabilities of large language models (LLMs) into small language models (SLMs) is essential for their practical application. The predominant approach, supervised fine-tuning (SFT), is an off-policy distillation method that suffers from poor out-of-domain (OOD) generalization because it rigidly aligns with static teacher trajectories. 
While reinforcement learning (RL) offers an alternative, the capacity limitations of SLMs pose a severe dilemma: sparse outcome rewards provide insufficient guidance, whereas strict trajectory matching imposes overly restrictive constraints. To bridge this capacity-driven gap, we propose \ours{}, an on-policy distillation framework that introduces a flexible yet process-aware reward structure. Instead of enforcing rigid replication, \ours{} uses the teacher's reference to guide tool-use behavior, balancing behavioral alignment with downstream performance. Extensive experiments on controlled executable-tool benchmarks demonstrate that \ours{} improves OOD tool-use performance compared to SFT and strict RL baselines. Our findings suggest that within verifiable tool-use environments, flexible tool-use alignment offers a more effective approach than strict trajectory replication for developing adaptable small models.

\end{abstract}

\section{Introduction}

The integration of large language models (LLMs) with external tools, such as code interpreters and retrieval APIs, has enabled them as advanced agents capable of handling complex reasoning tasks~\cite{yao2023react, paranjape2023art, wang2024executable, singh2025agentic}. However, the high inference costs of these large-scale models limit their practical use. This challenge has motivated a line of research focused on smaller language models (SLMs) to transfer the tool-assisted problem-solving capabilities of larger models~\cite{gao2023pal, gou2024tora, qiu2025agentdistill}.

To date, knowledge distillation has largely relied on off-policy imitation through supervised fine-tuning (SFT), where a student model is trained to strictly replicate teacher-generated trajectories~\cite{liu2024mobilellm, kang2025distilling, lyu2025correction}. However, this rigid alignment with static token distributions stifles the exploratory autonomy of capacity-limited SLMs. Consequently, forcing exact imitation hinders adaptive tool use and severely limits generalization to out-of-domain (OOD) scenarios and to unseen tools~\cite{luo2025self, sun2025climbing, yin2025enhancing}.

To mitigate reliance on static data, reinforcement learning (RL) enables models to internalize tool-use logic through interaction with the tool environment~\cite{trung2024reft, wu2025generalization}. Unlike SFT, RL refines policies through trial and error, fostering strategies that generalize beyond predefined examples~\cite{yu2024steptool, singh2025agentic, wu2025recode}. However, designing reward structures for capacity-limited SLMs poses a dilemma: outcome-only rewards provide weak guidance, leading to inefficient exploration, while strict trajectory matching imposes an overly restrictive optimization burden given their limited capacity. As a result, SLMs struggle to learn effective policy updates and underlying tool-planning logic.

To address the inherent limitations of SLM adaptation, we introduce three research questions to guide RL-based distillation in optimizing the strategic tool-use capabilities and performance of SLMs:
\textbf{RQ1:} What primary failure modes characterize strategic misalignment between SLMs and LLMs during tool-integrated reasoning?
\textbf{RQ2:} How can we design a reward structure that fosters flexible tool-level alignment, helping SLMs acquire teacher-guided tool-selection behavior?
\textbf{RQ3:} Can this flexible, on-policy distillation effectively guide capacity-limited SLMs to achieve robust task performance and alignment with the teacher?

Our preliminary observations indicate a significant difference in tool-use strategies between LLMs and SLMs, which primarily stems from the capacity limitations of SLMs. This makes it difficult to provide proper reinforcement signals, balancing between being too uninformative and being overly strict. To address this, we propose \textbf{\ours{}} (\textbf{M}odel \textbf{E}nhancement by \textbf{N}on-rigid \textbf{T}eacher-\textbf{O}ptimized \textbf{R}ewards), an on-policy distillation that mitigates this exploration failure by providing a flexible, reward-driven alignment signal that bridges the strategic gap between the teacher and the student.

In our experiments across multiple tool-use benchmarks, we demonstrate that \ours{} significantly outperforms both SFT and RL baselines. Notably, our approach improves out-of-domain tool-use performance, effectively narrowing the performance gap between 1.5B--8B models and their 235B teacher. These results confirm that replacing rigid trajectory memorization with flexible, teacher-guided tool-use exploration improves SLM generalization to unseen tool-use settings.\footnote{Our code is publicly available: \url{https://github.com/choics2623/MENTOR-RL/}}
To summarize, our key contributions are:
\begin{itemize}[leftmargin=*,topsep=-2px,partopsep=0px]
\item We empirically identify the fragile tool-use behavior of capacity-limited SLMs as a primary bottleneck, creating a severe reward dilemma between providing uninformative outcome signals and imposing overly restrictive constraints.
\item We propose \ours{}, a reinforcement learning framework that mitigates this dilemma by deploying a flexible yet process-aware reward structure. Our method provides moderated guidance for acquiring teacher-guided tool-use patterns while accommodating the SLM's limited capacity.
\item Our experiments show that \ours{} improves out-of-domain tool-use performance across multiple benchmarks. Replacing rigid trajectory imitation with flexible tool-use distillation reduces the performance gap between SLMs and LLMs more effectively than SFT and other RL approaches.
\end{itemize}

\section{Related Work}

\subsection{Tool-Integrated Reasoning of LLMs}
Tool-integrated reasoning (TIR) enhances LLM reasoning by integrating external tools, such as code interpreters and retrieval APIs. These approaches range from prompting-based integration~\cite{gao2023pal, paranjape2023art, huang2024planning} to explicit fine-tuning for tool invocation~\cite{schick2023toolformer, kong2023tptu, qian2024toolink}. Recognizing that effective tool use is an inherently strategic process, recent studies have explored learning optimal invocation patterns through reinforcement learning~\cite{yu2024steptool, feng2025retool, qian2025toolrl}. However, while large-scale models have made significant progress, extending robust tool-use policies to SLMs requires addressing distinct capacity-driven bottlenecks~\cite{kang2025distilling}. To address these constraints, we investigate the underlying failure modes in SLM exploration and propose \ours{}, which facilitates flexible strategic alignment.

\subsection{Distillation from LLMs to SLMs}
To transfer advanced capabilities to SLMs, knowledge distillation typically employs supervised fine-tuning (SFT) to clone teacher-generated trajectories~\cite{liu2024mobilellm, kang2025distilling, song2025rationale, lyu2025correction}. Despite its prevalence, SFT-based distillation faces limitations in out-of-domain generalization, as static datasets cannot cover unseen reasoning paths~\cite{sun2025climbing, yin2025enhancing}. A critical consequence is that student models tend toward \textbf{imitation} by matching the token distribution of trajectories without \textbf{internalizing} the underlying strategic logic~\cite{kandpal2023large, dai2024beyond, li2025llms}. In contrast to these static off-policy methods, \ours{} formulates an on-policy distillation framework. Rather than enforcing token-level logit matching on student rollouts, \ours{} leverages the teacher's trajectory as a strategic reward reference to guide the student's on-policy exploration.

\subsection{Reinforcement Learning for Tool-Use}
Reinforcement learning~\cite{kaelbling1996reinforcement} offers a dynamic alternative to SFT by enabling models to refine their policies through environmental interaction. Recent advances, such as group relative policy optimization (GRPO), underscore that policy optimization can effectively foster self-corrective reasoning behaviors~\cite{shao2024deepseekmath}, with the precise design of the reward function being a decisive factor.

This design choice becomes particularly critical when considering model scale. While highly capable LLMs can discover effective policies from sparse outcome rewards~\cite{yu2024steptool, feng2025retool, wu2025recode}, capacity-limited SLMs are notably less efficient at exploration~\cite{wei2022emergent, xiong2024watch}. Existing frameworks for larger models have introduced rewards on final outcomes~\cite{feng2025retool, singh2025agentic} or enforced strict alignment with ground-truth trajectories~\cite{qian2025toolrl}. However, these frameworks provide insufficient signals to induce tool use or impose an overly restrictive optimization burden. In contrast, our work leverages teacher trajectories to construct a flexible, process-aware reward signal that provides moderated yet adaptable guidance, bridging this critical optimization gap.

\section{\label{sec:observation}Observation: Limitations of SLM Exploration in TIR (RQ1)}

\begin{figure}[t]
    \centering
    \begin{subfigure}[t]{0.49\columnwidth}
        \centering
        \includegraphics[width=\textwidth]{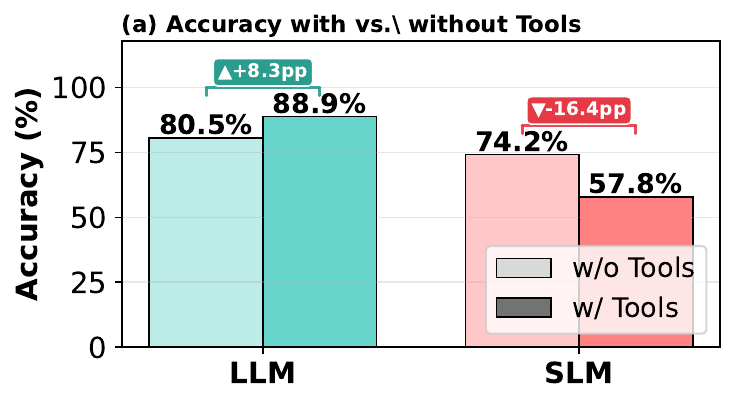}
        \vspace{-0.1in}
    \end{subfigure}\hfill
    \begin{subfigure}[t]{0.49\columnwidth}
        \centering
        \includegraphics[width=\textwidth]{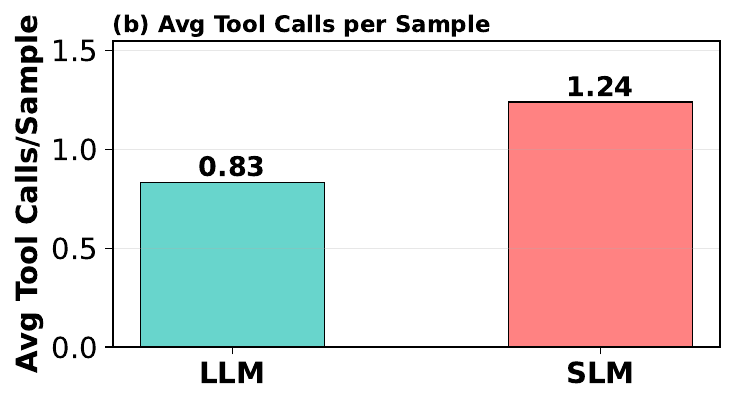}
        \vspace{-0.1in}
    \end{subfigure}

    \vspace{1mm}
    \begin{subfigure}[t]{\columnwidth}
        \centering
        \includegraphics[width=0.95\textwidth]{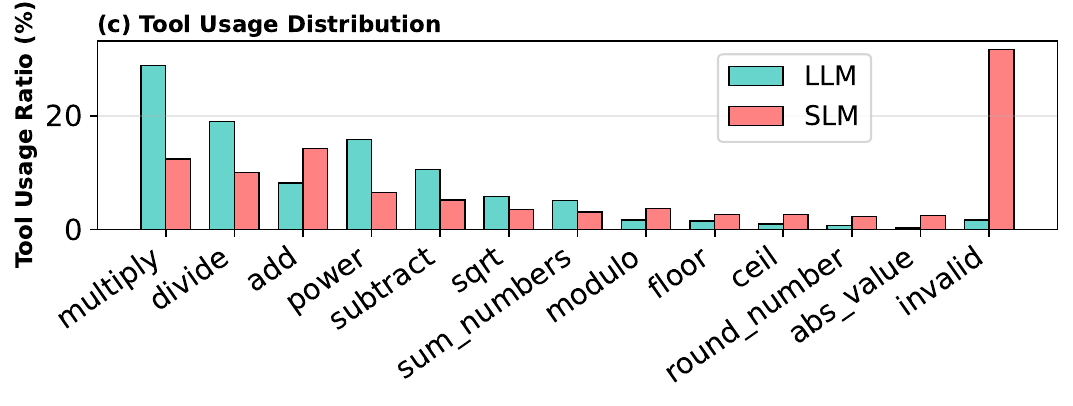}
        \vspace{-0.1in}
    \end{subfigure}
    \vspace{-0.3in}
    \caption{\label{fig:observation}
        Comparative analysis of tool-use behavior between LLM (Qwen3-235B)
        and SLM (Qwen3-8B).}
    \vspace{-0.15in}
\end{figure}

\begin{figure}[t]
    \centering
    \begin{subfigure}[t]{\columnwidth}
        \centering
        \includegraphics[width=0.8\textwidth]{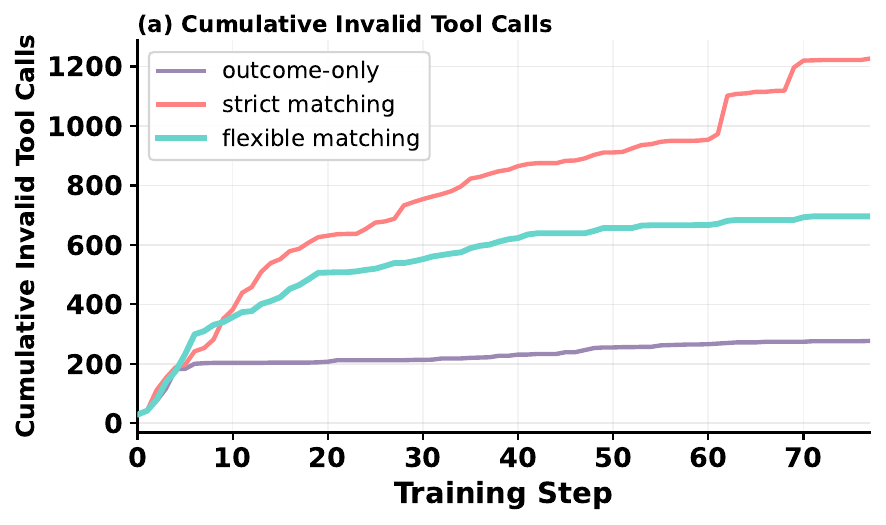}
        \vspace{-0.2in}
        \label{fig:abl_cum_invalid}
    \end{subfigure}

    \vspace{1mm}
    \begin{subfigure}[t]{\columnwidth}
        \centering
        \includegraphics[width=0.8\textwidth]{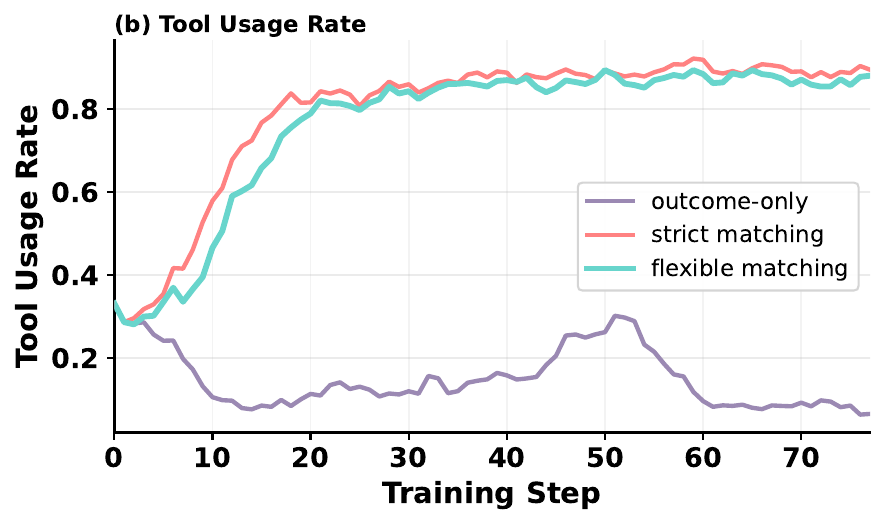}
        \vspace{-0.2in}
        \label{fig:abl_usage_rate}
    \end{subfigure}
    \vspace{-0.3in}
    \caption{\label{fig:observation_learning}Comparison of various reward settings, including outcome-only, strict matching, and flexible matching (\ours{}) over training steps.}
\end{figure}

To investigate the root causes of the performance gap between SLMs and LLMs, we conduct a comparative analysis of their exploration behaviors during TIR. Specifically, we compare the tool-use patterns of an LLM (Qwen3-235B) and an SLM (Qwen3-8B) (\autoref{fig:observation}) and analyze the learning curves of the SLM under various reward settings (\autoref{fig:observation_learning}). Our analysis demonstrates that this inherent capacity gap manifests as distinct behavioral bottlenecks, posing critical optimization challenges in reinforcement learning.

%

\paragraph{Performance Disparity Despite Higher Tool-Use Frequency.}
Our results show that the SLM achieves a lower success rate than the teacher (\autoref{fig:observation}a), even though it invokes tools at a higher frequency (\autoref{fig:observation}b). This discrepancy suggests that the performance gap stems not from a lack of tool engagement but from the SLM's inability to employ tools effectively. Consequently, bridging this gap requires the SLM to move beyond mere tool invocation and internalize the teacher's underlying strategic logic.

\paragraph{Divergence in Tool Selection Patterns.}
This performance disparity is further evidenced by their distinct tool-selection distributions (\autoref{fig:observation}c). Even when facing identical problems, the SLM exhibits a markedly different invocation pattern compared to the LLM. While the larger model strategically focuses on complex operational tools, the student model relies heavily on simpler or redundant functions and frequently suffers from execution failures. This clear divergence indicates a fundamental gap in strategic planning capabilities, underscoring that improving TIR in SLMs requires aligning their planning logic with the advanced strategy of the teacher. Example cases contrasting LLM and SLM tool selections, along with the vulnerability from invalid tool calls, are provided in Appendix~\ref{appen:example_case}.

\begin{figure*}[t]
    \begin{center}
    \includegraphics[width=0.85\linewidth]{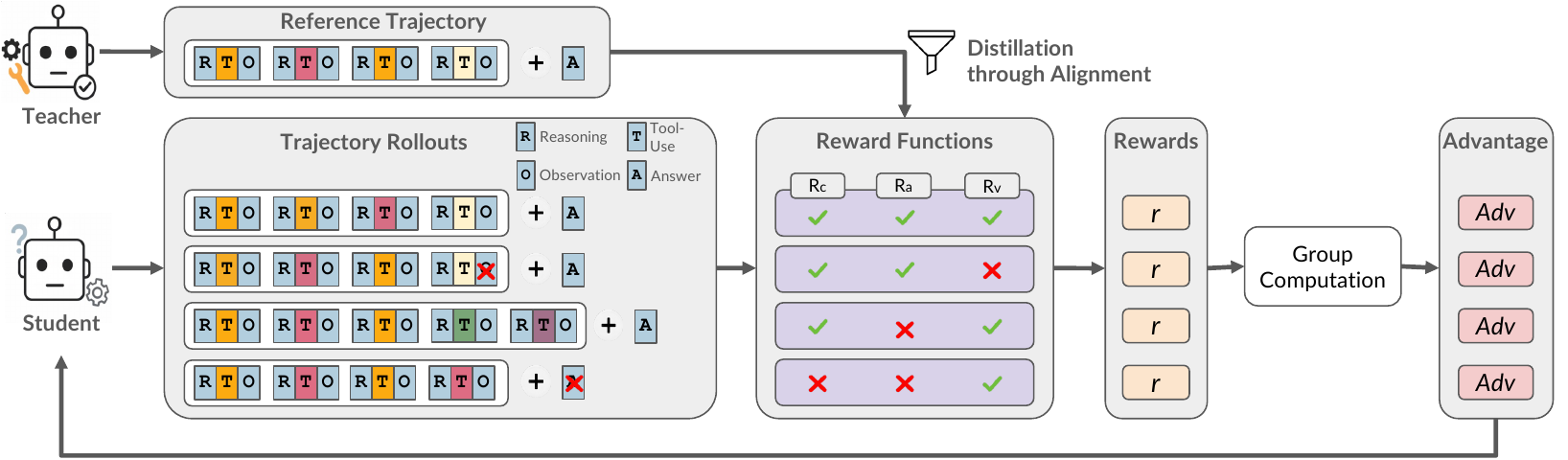}
    \end{center}
    \vspace{-0.15in}
    \caption{ Overview of the \ours{} training framework. A problem-solving trajectory ($\tau$) consists of a sequence of Reasoning (R), Tool-Use (T), and Observation (O), and a final Answer (A). Each student rollout is evaluated by a set of reward models, which generate a reward signal by aligning the student's actions against the teacher's reference trajectory.}
    \label{fig:framework_overview}
    \vspace{-0.1in}
\end{figure*}

\paragraph{The Reward Dilemma in SLM Exploration.}
The divergence in tool-selection patterns and high rates of execution failure create a severe initial exploration bottleneck in reinforcement learning. To investigate this, we analyze the learning curves of SLMs across various reward settings, evaluating overall tool usage rates and the cumulative number of invalid tool calls. As shown, deploying a sparse, outcome-only reward provides an insufficient reinforcement signal, causing the SLM to avoid using tools (\autoref{fig:observation_learning}a). Conversely, enforcing strict trajectory matching encourages tool invocation, but the rigid sequence constraint is overly restrictive. As training continues, this strict guidance leads to a steady increase in the total number of invalid tool calls (\autoref{fig:observation_learning}b). This trade-off demonstrates that overcoming exploration failure requires a balanced reward structure that both induces high tool engagement and suppresses invalid tool calls.


\section{Method: Designing Rewards for Tool-Use Alignment (RQ2)}
To address the exploration bottlenecks and optimization challenges identified in Section~\ref{sec:observation}, we introduce \ours{}, a reinforcement learning framework that provides balanced, process-aware rewards. Rather than relying solely on sparse outcome rewards or enforcing strict trajectory imitation, our method uses teacher trajectories to provide moderated guidance, enabling the student to acquire high-level tool-selection behavior guided by the teacher.

\subsection{\label{sec:method_overview}Methodology Overview}
The overall process of \ours{} is illustrated in \autoref{fig:framework_overview}.
Given a training dataset $\mathcal{D}=\{(x_i,y_i)\}^N_{i=1}$, where $x_i$ represents a question and $y_i$ is its corresponding ground-truth answer, the teacher model generates an output $O^{(t)}$ for each question $x$. This output consists of a final answer $\hat{y}^{(t)}$ and a trajectory $\tau^{(t)}$:
\begin{equation}
\small
    O^{(t)}=(\tau^{(t)}, \hat{y}^{(t)}) \sim \pi_\text{teacher}(\cdot |x,I),
\end{equation}
\begin{equation}
\small
    \tau^{(t)} = \langle(r_1,t_1,o_1), \dots, (r_{L_\tau},t_{L_\tau},o_{L_\tau})\rangle,
\end{equation} 
where $I$ is an instruction prompt and the trajectory $\tau^{(t)}$ is a sequence of reasoning ($r$), tool-use ($t$), and observation ($o$) steps. 
Subsequently, the student model performs on-policy exploration by generating a group of rollouts, each denoted as $O^{(s)}$. By evaluating these rollouts against the teacher's reference trajectory, \ours{} provides a balanced guidance signal that enables the student to acquire teacher-guided tool-use patterns. We optimize the student's policy using the GRPO framework under this reward-driven formulation. Detailed procedures and the objective function are provided in Appendix~\ref{appen:method_detail}.

\subsection{Flexible Reward Design}
Our reward functions are designed to mitigate the two primary trade-offs identified in Section~\ref{sec:observation}: tool avoidance and continuous invalid calls. We define a composite reward signal $R(O^{(s)},O^{(t)})$ that integrates three key components: correctness ($R_c$), strategic alignment ($R_a$), and tool validation ($R_v$):
\begin{equation}\label{eq:composite-reward}R(O^{(s)},O^{(t)}) = w_c R_c + w_a R_a + w_v R_v\end{equation}
where $w_c, w_a,$ and $w_v$ are hyperparameters that balance the contributions of each reward component. We provide a detailed analysis of these weights and their optimization in Appendix~\ref{appen:weights}.

\paragraph{Design Background.}
While prior frameworks have explored various reward structures, they typically frame tool-use optimization either as a sparse outcome-matching problem~\cite{feng2025retool,singh2025agentic} or as a rigid sequence-imitation task~\cite{qian2025toolrl}. However, these empirical trade-offs suggest that neither approach is well suited to capacity-limited SLMs, which require intermediate guidance without being forced to replicate a teacher's trajectory exactly.

To bridge this gap, \ours{} shifts the optimization objective from literal sequence imitation to flexible, strategic alignment. Rather than penalizing minor structural variations or individual token mismatches, our set-based design transfers the teacher's core tool-selection strategy while allowing the student to follow alternative valid reasoning paths. This approach provides a balanced, process-aware learning environment, delivering sufficient reinforcement signals to maintain active engagement with the tool while granting the student autonomy to explore valid reasoning paths.

\paragraph{Final Correctness via $R_c$.} Consistent with prior TIR frameworks~\cite{feng2025retool, singh2025agentic, qian2025toolrl}, we include the \textit{Correctness Reward} ($R_c$) to assess whether the student's final answer $\hat{y}^{(s)}$ matches the teacher's outcome $\hat{y}^{(t)}$. For robust distillation, we use only reference trajectories in which the teacher’s answer is consistent with the ground truth.
\begin{equation}
\small
    R_{\text{c}} =
    \begin{cases}
    1 & \text{if } \hat{y}^{(s)} =  \hat{y}^{(t)} \\[6pt]
    0 & \text{otherwise}
    \end{cases}
\end{equation}

\paragraph{Internalizing Strategic Planning via $R_a$.} To focus on strategic alignment, we introduce the \textit{Teacher-Alignment Reward} ($R_a$). Rather than enforcing strict adherence to sequence formats, $R_a$ prioritizes strategic flexibility by rewarding the student for using the same set of essential tools as the teacher. This is defined as:
\begin{equation}
\small
    R_a = \begin{cases} 1 & \text{if } \{t \mid t \in \tau^{(s)}\} = \{t \mid t \in \tau^{(t)}\} \\[6pt] 0 & \text{otherwise} \end{cases}
    \label{eq:teacher_alignment}
\end{equation}
As shown in \autoref{fig:framework_overview}, where the first student rollout achieves alignment despite a different tool execution order, this set-matching approach allows the student to explore alternative valid sequences without penalty. A detailed comparison with alternative approaches is provided in Appendix~\ref{appen:set_base}.

\paragraph{Addressing Invalid Tool Calls via $R_v$.} Since we do not enforce a strict order of tool use, the student model must ensure that all invoked tools are executed validly. To maintain the viability of execution, the \textit{Tool Validation Reward} ($R_v$) penalizes non-executable or malformed tool calls in the student's trajectory $\tau^{(s)}$. This reward is granted only if every tool call in the student's trajectory executes successfully without raising an error:
\begin{equation}
\small
    R_{\text{v}} =
    \begin{cases}
    1 & \forall o_i^{(s)} \in \tau^{(s)} \text{ are valid}\\
    0 & \text{otherwise}
    \end{cases}
\end{equation}

\section{Experiments: Effectiveness of Rewards for Tool-Use Alignment (RQ3)}

\subsection{Experimental Setup \label{sec:experiment-setup}}
\paragraph{Training Domain Selection.}
We adopt mathematical reasoning as a controlled training domain for studying reward design, as it provides well-defined success criteria and structured tool-use trajectories. Its inherent difficulty gradient drives robust exploration beyond simple policies, mitigating the risk of suboptimal convergence~\cite{zhou2023solving, wang2024mathcoder, acereason, luo2025wizardmath}. This choice is also supported by prior work showing that mathematical ability is a transferable skill that provides a strong foundation for generalizable problem-solving~\cite{wang2025octothinker, huang2025r}. We use the AceReason-Math dataset for training~\cite{acereason}. Details of our training dataset are provided in Appendix~\ref{appen:training_data}.

\paragraph{Models and Tools.}
Our teacher model is Qwen3-235B-Thinking, chosen for its strong tool-use capabilities. We use four student models to evaluate our method across different scales: Qwen3 (8B and 1.7B) and Qwen2.5 (7B and 1.5B). To augment the models with tool-use capabilities, we implement a sandbox to execute Python code on a remote server. While our primary experiments focus on the Qwen family, analyses demonstrating the extensibility of our framework to other model families are provided in Appendix~\ref{appen:extensibility}. Details on model versions, tools, and further implementation details are provided in Appendices~\ref{appen:models}, \ref{appen:math_tool}, and \ref{appen:implementation}.

\renewcommand\cellalign{tl} 
\newcommand{\gcell}[3]{\makecell[l]{\textbf{#1}\\ \textit{#2}\\ #3}}
\newcommand{\gcellBaseline}[2]{\makecell[l]{\textbf{#1}\\ #2}}

\begin{table*}[t!]
\centering
\scriptsize
\begin{adjustbox}{max width=\textwidth, scale = 0.8}
\begin{tabular}{ll|ccccccc}
\toprule
\multirow{2}{*}{\textbf{Model}} & \multirow{2}{*}{\textbf{Method}} 
  & \multicolumn{7}{c}{\textbf{Math (Acc)}}  \\
\cmidrule(lr){3-9} 
 & & Math-Forge & Omni-MATH & aime24 & aime25 & amc23 & minervamath & Overall  \\
\midrule
\multirow{1}{*}{\textbf{Qwen 3}}
  & 235B-Vanilla & 68.00 & 26.00 & 46.67 & 40.00 & 85.00 & 46.69 & 52.06 \\
\midrule
\multirow{8}{*}{\textbf{Qwen 2.5}} 
  & 1.5B-Vanilla & 5.18 \scriptsize{($\pm$0.11)} & 1.23 \scriptsize{($\pm$0.18)} & 0.00 \scriptsize{($\pm$0.00)} & 0.00 \scriptsize{($\pm$0.00)} & 2.50 \scriptsize{($\pm$2.04)} & 1.29 \scriptsize{($\pm$0.56)}& 1.70 \\
  & 1.5B-SFT     & 5.79 \scriptsize{($\pm$0.09)} & 1.62 \scriptsize{($\pm$0.23)} & 1.00 \scriptsize{($\pm$2.25)} & 1.33 \scriptsize{($\pm$1.72)} & 5.75 \scriptsize{($\pm$2.90)} & 3.53 \scriptsize{($\pm$0.43)} & 3.17 \\
  & 1.5B-Outcome-based  & 18.38 \scriptsize{($\pm$0.09)} & 3.46 \scriptsize{($\pm$0.18)} & 2.67 \scriptsize{($\pm$3.06)} & 3.00 \scriptsize{($\pm$2.92)} & 8.50* \scriptsize{($\pm$2.93)} & 4.12 \scriptsize{($\pm$0.34)} & 6.69 \\
  & 1.5B-Strict RL  & 18.62 \scriptsize{($\pm$0.10)} & 3.78 \scriptsize{($\pm$0.21)} & 2.33 \scriptsize{($\pm$2.85)} & 3.33 \scriptsize{($\pm$1.95)} & 9.25*
  \scriptsize{($\pm$2.78)} & 4.38 \scriptsize{($\pm$0.38)} & 6.95 \\
  & 1.5B-\ours{}   & \textbf{18.84}* \scriptsize{($\pm$0.13)} & \textbf{5.64}* \scriptsize{($\pm$0.28)} & \textbf{10.00}* \scriptsize{($\pm$3.52)} & \textbf{6.67}* \scriptsize{($\pm$2.72)} & \textbf{9.75}* \scriptsize{($\pm$2.49)} & \textbf{8.42}* \scriptsize{($\pm$0.44)} & \textbf{9.89} \\
  \cmidrule(lr){2-9}
  & 7B-Vanilla   & 36.87 \scriptsize{($\pm$1.04)} & 8.60 \scriptsize{($\pm$1.28)} & 6.66 \scriptsize{($\pm$2.22)} & 5.33* \scriptsize{($\pm$2.81)} & 44.00 \scriptsize{($\pm$3.16)} & 26.48 \scriptsize{($\pm$0.96)} & 21.32 \\
  & 7B-SFT       & 37.64 \scriptsize{($\pm$1.25)} & 11.14 \scriptsize{($\pm$1.99)} & 9.40* \scriptsize{($\pm$2.65)} & \textbf{7.33}* \scriptsize{($\pm$3.44)} & 43.00 \scriptsize{($\pm$4.53)} & 26.38 \scriptsize{($\pm$2.64)} & 22.48 \\
  & 7B-Outcome-based    & 50.25 \scriptsize{($\pm$1.32)} & 12.94 \scriptsize{($\pm$1.21)} & 12.00* \scriptsize{($\pm$3.58)} & 6.66* \scriptsize{($\pm$2.22)} & 45.50 \scriptsize{($\pm$3.07)} & 29.40 \scriptsize{($\pm$1.70)} & 26.13 \\
  & 7B-Strict RL    & 50.85 \scriptsize{($\pm$1.41)} & 13.42 \scriptsize{($\pm$1.18)} & 11.67* \scriptsize{($\pm$3.50)} & 7.00* \scriptsize{($\pm$2.85)} & 46.50
  \scriptsize{($\pm$3.20)} & 30.40 \scriptsize{($\pm$1.85)} & 26.64 \\
  & 7B-\ours{}     & \textbf{55.42}* \scriptsize{($\pm$1.96)} & \textbf{14.72}* \scriptsize{($\pm$2.21)} & \textbf{12.33}* \scriptsize{($\pm$3.52)} & \textbf{7.33}* \scriptsize{($\pm$2.63)} & \textbf{50.10}* \scriptsize{($\pm$2.70)} & \textbf{31.86}* \scriptsize{($\pm$1.71)} & \textbf{28.63}\\
\midrule
\multirow{8}{*}{\textbf{Qwen 3}} 
  & 1.7B-Vanilla & 58.70 \scriptsize{($\pm$0.11)} & 16.28 \scriptsize{($\pm$0.21)} & 25.36 \scriptsize{($\pm$2.32)} & 13.01 \scriptsize{($\pm$2.44)} & 58.00 \scriptsize{($\pm$4.22)} & {28.85} \scriptsize{($\pm$0.47)} & 33.37 \\
  & 1.7B-SFT & 59.99 \scriptsize{($\pm$0.12)} & 16.65 \scriptsize{($\pm$0.21)} & 31.00 \scriptsize{($\pm$2.25)} & {16.02} \scriptsize{($\pm$2.10)} & 63.75 \scriptsize{($\pm$3.17)} & 30.60 \scriptsize{($\pm$0.52)} & 36.34 \\
  & 1.7B-Outcome-based & 60.18 \scriptsize{($\pm$0.07)} & 17.78 \scriptsize{($\pm$0.28)} & 34.34 \scriptsize{($\pm$3.13)} & 20.00 \scriptsize{($\pm$2.22)} & {67.75} \scriptsize{($\pm$2.75)} & 31.67 \scriptsize{($\pm$0.48)} & 38.62 \\
  & 1.7B-Strict RL  & 60.42 \scriptsize{($\pm$0.09)} & 18.34 \scriptsize{($\pm$0.31)} & 33.67 \scriptsize{($\pm$3.35)} & 20.67 \scriptsize{($\pm$2.34)} & 68.50
  \scriptsize{($\pm$2.95)} & 32.05 \scriptsize{($\pm$0.45)} & 38.94 \\
  & 1.7B-\ours{} & \textbf{60.66}* \scriptsize{($\pm$0.11)} & \textbf{20.74}* \scriptsize{($\pm$0.22)} & \textbf{36.67}* \scriptsize{($\pm$2.50)} & \textbf{24.00}* \scriptsize{($\pm$4.08)} & \textbf{71.00}* \scriptsize{($\pm$3.94)} & \textbf{32.30}* \scriptsize{($\pm$0.38)} & \textbf{40.90} \\
  \cmidrule(lr){2-9}
  & 8B-Vanilla & 65.00 \scriptsize{($\pm$0.12)} & 19.46 \scriptsize{($\pm$0.24)} & 29.32 \scriptsize{($\pm$2.64)} & 20.65 \scriptsize{($\pm$3.44)} & 57.75 \scriptsize{($\pm$2.49)} & 42.55 \scriptsize{($\pm$0.58)} & 39.12 \\
  & 8B-SFT & 65.35 \scriptsize{($\pm$0.07)} & 20.45 \scriptsize{($\pm$0.21)} & {31.65} \scriptsize{($\pm$3.23)} & 25.65 \scriptsize{($\pm$4.17)} & 62.50 \scriptsize{($\pm$2.89)} & 43.22 \scriptsize{($\pm$0.31)} & 41.47 \\
  & 8B-Outcome-based & {65.48} \scriptsize{($\pm$0.04)} & 22.03 \scriptsize{($\pm$0.26)} & 35.33 \scriptsize{($\pm$3.22)} & {30.33} \scriptsize{($\pm$3.99)} & 75.50 \scriptsize{($\pm$2.58)} & 43.49 \scriptsize{($\pm$0.25)} & 45.19 \\
  & 8B-Strict RL    & 65.76 \scriptsize{($\pm$0.06)} & 22.48 \scriptsize{($\pm$0.27)} & 35.00 \scriptsize{($\pm$3.40)} & 31.00 \scriptsize{($\pm$4.10)} & 76.50
  \scriptsize{($\pm$2.65)} & 43.85 \scriptsize{($\pm$0.32)} & 45.77 \\
  & 8B-\ours{}  & \textbf{66.50}* \scriptsize{($\pm$0.11)} & \textbf{24.10}* \scriptsize{($\pm$0.25)} & \textbf{39.00}* \scriptsize{($\pm$3.53)} & \textbf{37.00}* \scriptsize{($\pm$3.31)} & \textbf{79.00}* \scriptsize{($\pm$2.69)} & \textbf{43.94}* \scriptsize{($\pm$0.44)} & \textbf{48.26} \\
\bottomrule
\end{tabular}
\end{adjustbox}

\vspace{0.1em} 

\begin{adjustbox}{max width=\textwidth, scale = 0.9}
\begin{tabular}{ll|ccccccccc}
\toprule
\multirow{2}{*}{\textbf{Model}} & \multirow{2}{*}{\textbf{Method}} 
  & \multicolumn{5}{c}{\textbf{BFCL-v4 (Acc)}} 
  & \multicolumn{4}{c}{\textbf{RAG (EM)}}  \\
\cmidrule(lr){3-7} \cmidrule(lr){8-11} 
 & & Non-Live & Multi-Turn & Live & Agentic & Overall 
   & Bamboogle & 2WikiMultiHopQA & HotpotQA & Overall \\
\midrule
\multirow{1}{*}{\textbf{Qwen 3}}
 & 235B-Vanilla
  & 87.62 & 51.88 & 82.68 & 18.83 & 50.13 
  & 41.60 & 42.50 & 34.60 & 39.57 \\
\midrule
\multirow{8}{*}{\textbf{Qwen 2.5}} 
  & 1.5B-Vanilla 
  & 68.88 \scriptsize{($\pm$1.15)} & 1.09 \scriptsize{($\pm$0.34)} & 58.99 \scriptsize{($\pm$0.54)} & 2.21 \scriptsize{($\pm$0.44)} & 24.00
  & 0.30 \scriptsize{($\pm$0.31)} & 3.20 \scriptsize{($\pm$0.55)} & 1.47 \scriptsize{($\pm$0.60)} & 1.66 \\
  & 1.5B-SFT 
  & {69.96} \scriptsize{($\pm$1.20)} & 1.58 \scriptsize{($\pm$0.42)} & 60.72 \scriptsize{($\pm$0.36)} & 2.65 \scriptsize{($\pm$0.29)} & 24.59
  & 0.85 \scriptsize{($\pm$0.53)} & 3.91 \scriptsize{($\pm$0.47)} & 2.52 \scriptsize{($\pm$0.57)} & 2.43 \\
  & 1.5B-Outcome-based 
  & 71.38 \scriptsize{($\pm$0.63)} & 1.98 \scriptsize{($\pm$0.50)} & 61.46 \scriptsize{($\pm$0.45)} & 3.09 \scriptsize{($\pm$0.39)} & 25.12 
  & 6.80 \scriptsize{($\pm$0.49)} & 10.19 \scriptsize{($\pm$0.41)} & 5.79 \scriptsize{($\pm$0.66)} & 7.59 \\
    & 1.5B-Strict RL
    & 71.75 \scriptsize{($\pm$0.59)} & 1.92 \scriptsize{($\pm$0.46)} & 61.62 \scriptsize{($\pm$0.42)} & 2.78 \scriptsize{($\pm$0.36)} & 24.85
    & 6.92 \scriptsize{($\pm$0.52)} & 6.85 \scriptsize{($\pm$0.48)} & 5.42 \scriptsize{($\pm$0.62)} & 6.39 \\
  & 1.5B-\ours{} 
  & \textbf{72.66}* \scriptsize{($\pm$0.62)} & \textbf{2.35}* \scriptsize{($\pm$0.41)} & \textbf{62.03}* \scriptsize{($\pm$0.39)} & \textbf{3.83}* \scriptsize{($\pm$0.40)} &  \textbf{25.70}
  & \textbf{7.36}* \scriptsize{($\pm$0.55)} & \textbf{11.31}* \scriptsize{($\pm$0.54)} & \textbf{6.50}* \scriptsize{($\pm$0.40)} & \textbf{8.39} \\
  \cmidrule(lr){2-11}
  & 7B-Vanilla 
  & 72.04 \scriptsize{($\pm$0.33)} & 3.90 \scriptsize{($\pm$1.43)} & 63.78 \scriptsize{($\pm$0.91)} & 4.09 \scriptsize{($\pm$1.86)} & 26.38 & 14.88 \scriptsize{($\pm$2.48)} & 14.20 \scriptsize{($\pm$0.58)} & 13.08 \scriptsize{($\pm$1.00)} & 14.05 \\
  & 7B-SFT   
  & 72.05 \scriptsize{($\pm$0.32)} & 4,02 \scriptsize{($\pm$1.08)} & 63.79 \scriptsize{($\pm$0.79)} & 4.27 \scriptsize{($\pm$1.76)} & 26.50 & 16.48 \scriptsize{($\pm$1.74)} & 13.87 \scriptsize{($\pm$0.53)} & 13.36 \scriptsize{($\pm$0.70)} & 14.57 \\
  & 7B-Outcome-based
  & 82.13 \scriptsize{($\pm$0.09)} & 14.14* \scriptsize{($\pm$1.69)} & 72.22 \scriptsize{($\pm$0.30)} & 8.15 \scriptsize{($\pm$1.15)} & 32.94 & 22.04 \scriptsize{($\pm$1.06)} & 17.57 \scriptsize{($\pm$0.55)} & 17.07 \scriptsize{($\pm$0.65)} & 18.89 \\
    & 7B-Strict RL
    & 82.65 \scriptsize{($\pm$0.14)} & 14.22* \scriptsize{($\pm$1.62)} & 72.62 \scriptsize{($\pm$0.34)} & 7.85 \scriptsize{($\pm$1.18)} & 33.15
    & 23.25 \scriptsize{($\pm$1.10)} & 17.20 \scriptsize{($\pm$0.58)} & 16.85 \scriptsize{($\pm$0.68)} & 19.10 \\
  & 7B-\ours{} & \textbf{83.35}* \scriptsize{($\pm$0.21}) & \textbf{14.27}* \scriptsize{($\pm$1.74)} & \textbf{72.81}* \scriptsize{($\pm$0.43)} & \textbf{9.30}* \scriptsize{($\pm$1.34)} & \textbf{33.52} & \textbf{23.56}* \scriptsize{($\pm$1.24)} & \textbf{19.39}* \scriptsize{($\pm$0.40)} & \textbf{19.24}* \scriptsize{($\pm$0.56)} & \textbf{20.73} \\
\midrule
\multirow{8}{*}{\textbf{Qwen 3}} 
  & 1.7B-Vanilla 
  & 80.89 \scriptsize{($\pm$0.70)} & 10.47 \scriptsize{($\pm$0.52)} & 69.90 \scriptsize{($\pm$0.22)} & 3.98 \scriptsize{($\pm$0.30)} & 29.81 
  & 14.61 \scriptsize{($\pm$0.53)} & 18.87 \scriptsize{($\pm$0.31)} & 16.11 \scriptsize{($\pm$0.36)} & 16.53 \\
  & 1.7B-SFT 
  & 81.55 \scriptsize{($\pm$0.50)} & 10.45 \scriptsize{($\pm$0.46)} & {70.36} \scriptsize{($\pm$0.41)} & 4.51 \scriptsize{($\pm$0.39)} & 30.13 
  & 15.95 \scriptsize{($\pm$0.54)} & 19.69 \scriptsize{($\pm$0.51)} & 17.00 \scriptsize{($\pm$0.43)} & 17.45 \\
  & 1.7B-Outcome-based
  & 83.00 \scriptsize{($\pm$0.31)} & 11.50 \scriptsize{($\pm$0.38)} & {70.77} \scriptsize{($\pm$0.31)} & 4.73 \scriptsize{($\pm$0.46)} & 30.65 
  & 16.83 \scriptsize{($\pm$0.55)} & 20.36 \scriptsize{($\pm$0.53)} & {17.35} \scriptsize{($\pm$0.29)}  & 18.18 \\
    & 1.7B-Strict RL
    & 83.55 \scriptsize{($\pm$0.34)} & 11.72 \scriptsize{($\pm$0.42)} & 70.62 \scriptsize{($\pm$0.32)} & 4.62 \scriptsize{($\pm$0.45)} & 30.95
    & 17.65 \scriptsize{($\pm$0.56)} & 20.05 \scriptsize{($\pm$0.58)} & 17.55 \scriptsize{($\pm$0.32)} & 18.42 \\
  & 1.7B-\ours{}  
  & \textbf{83.76}* \scriptsize{($\pm$0.35)} & \textbf{12.02}* \scriptsize{($\pm$0.46)} & \textbf{71.05}* \scriptsize{($\pm$0.34)} & \textbf{5.52}* \scriptsize{($\pm$0.60)} & \textbf{31.20} 
  & \textbf{18.28}* \scriptsize{($\pm$0.55)} & \textbf{21.37}* \scriptsize{($\pm$0.60)} & \textbf{18.03}* \scriptsize{($\pm$0.48)} & \textbf{19.23} \\
  \cmidrule(lr){2-11}
  & 8B-Vanilla 
  & 87.28 \scriptsize{($\pm$0.49)} & 35.51 \scriptsize{($\pm$0.47)} & 80.16 \scriptsize{($\pm$0.29)} & 9.89 \scriptsize{($\pm$0.33)} & 41.35
  & {32.13} \scriptsize{($\pm$0.64)} & 35.88 \scriptsize{($\pm$0.61)} & 27.47 \scriptsize{($\pm$0.61)} & 31.83 \\
  & 8B-SFT  
  & 87.91 \scriptsize{($\pm$0.48)} & 36.92 \scriptsize{($\pm$0.60)} & 80.59 \scriptsize{($\pm$0.33)} & 10.58 \scriptsize{($\pm$0.44)} & 42.16
  & {34.65} \scriptsize{($\pm$0.31)} & 37.18 \scriptsize{($\pm$0.36)} & 30.59 \scriptsize{($\pm$0.44)} &  34.14 \\
  & 8B-Outcome-based
  & 88.37 \scriptsize{($\pm$0.37)} & {38.06} \scriptsize{($\pm$0.29)} & {81.10} \scriptsize{($\pm$0.37)} & {11.03} \scriptsize{($\pm$0.34)} & {42.78}
  & 35.18 \scriptsize{($\pm$0.36)} & {37.95} \scriptsize{($\pm$0.30)} & {31.30} \scriptsize{($\pm$0.31)} & 34.81 \\
    & 8B-Strict RL
    & 88.62 \scriptsize{($\pm$0.38)} & 38.18 \scriptsize{($\pm$0.32)} & 81.32 \scriptsize{($\pm$0.40)} & 10.92 \scriptsize{($\pm$0.36)} & 42.95
    & 35.42 \scriptsize{($\pm$0.38)} & 37.45 \scriptsize{($\pm$0.32)} & 30.92 \scriptsize{($\pm$0.34)} & 34.66 \\
  & 8B-\ours{} 
  & \textbf{89.41}* \scriptsize{($\pm$0.56)} & \textbf{38.77}* \scriptsize{($\pm$0.60)} & \textbf{81.96}* \scriptsize{($\pm$0.52)} & \textbf{11.57}* \scriptsize{($\pm$0.42)} & \textbf{43.39}
  & \textbf{35.77}* \scriptsize{($\pm$0.36)} & \textbf{38.86}* \scriptsize{($\pm$0.56)} & \textbf{31.68}* \scriptsize{($\pm$0.36)} & \textbf{35.44} \\
\bottomrule
\end{tabular}
\end{adjustbox}
\vspace{-0.1in}
\caption{\label{tab:main_results} Main results comparing \ours{} against baselines across all evaluation benchmarks. The top table shows in-domain accuracy (\%) on mathematical reasoning tasks. The bottom table shows out-of-domain performance on BFCL-v4 (accuracy \%) and RAG (exact match \%). Results are reported as mean ($\pm$ standard deviation) over 10 runs. Overall scores are calculated differently: as a macro-average for MATH and RAG, and as an official weighted average for BFCL-v4. Values marked with an asterisk (*) denote statistical significance, based on Bonferroni-corrected pairwise test at $\alpha=0.05$.}
\end{table*}

\paragraph{Baselines.}
To evaluate our proposed framework, we compare \ours{} against four distinct baselines that represent different training approaches.
\textbf{1) Vanilla SLM:} The base instruction-tuned model, serving as a performance lower bound.
\textbf{2) SFT:} The predominant distillation approach, fine-tuning the student on static teacher-generated trajectories.
\textbf{3) Outcome-based RL:} To isolate the impact of our flexible reward design, we employ the same RL framework but utilize a simple reward signal based solely on the final outcome.
\textbf{4) Strict RL:} We also employ the same RL framework but rely on a strict sequential matching reward. Specifically, we re-implement and adapt the rigid trajectory-matching reward introduced in ToolRL~\cite{qian2025toolrl} to fit our experimental environment. Further implementation details are provided in Appendix~\ref{appen:baseline_implementation}.

\paragraph{Benchmarks.}
We evaluate our framework on a diverse set of in-domain and out-of-domain tasks to measure both task-specific performance and generalization ability. For \textbf{in-domain tasks}, our agent is evaluated on six mathematical reasoning benchmarks that provide a natural gradient of difficulty. For \textbf{out-of-domain tasks}, we assess zero-shot generalization by testing the agent on tools unseen during training. Specifically, we employ three benchmarks to evaluate retrieval-based QA and the BFCL v4 benchmark to test general tool-calling capabilities. Detailed descriptions of each dataset are provided in Appendix~\ref{appen:benchmark}.

\paragraph{Evaluation Metrics.}
We evaluate task performance using accuracy (Acc) and exact match (EM), where a prediction is considered correct only if it exactly matches one of the ground-truth answers after normalization.
To assess how effectively the student internalizes the teacher's advanced logic, we quantify policy alignment with an \textbf{alignment score (AS)} derived from the Jensen-Shannon divergence. This score measures the divergence between a model's tool-usage distribution and the teacher's reference distribution shown in \autoref{fig:observation}.
Further details are in Appendix~\ref{appen:metric}.

\subsection{Experimental Results}
The main results, presented in \autoref{tab:main_results}, demonstrate that our proposed framework, \ours{}, outperforms the baselines.

\paragraph{Distillation Enables Effective Tool Use.}
For in-domain tasks, both SFT and RL-based distillation improve tool-use effectiveness, yielding performance gains over vanilla baselines. This confirms that transferring the teacher's tool-calling capabilities serves as an effective strategy for enhancing SLM performance.

\begin{figure}[t]
    \centering
    \includegraphics[width=0.8\linewidth]{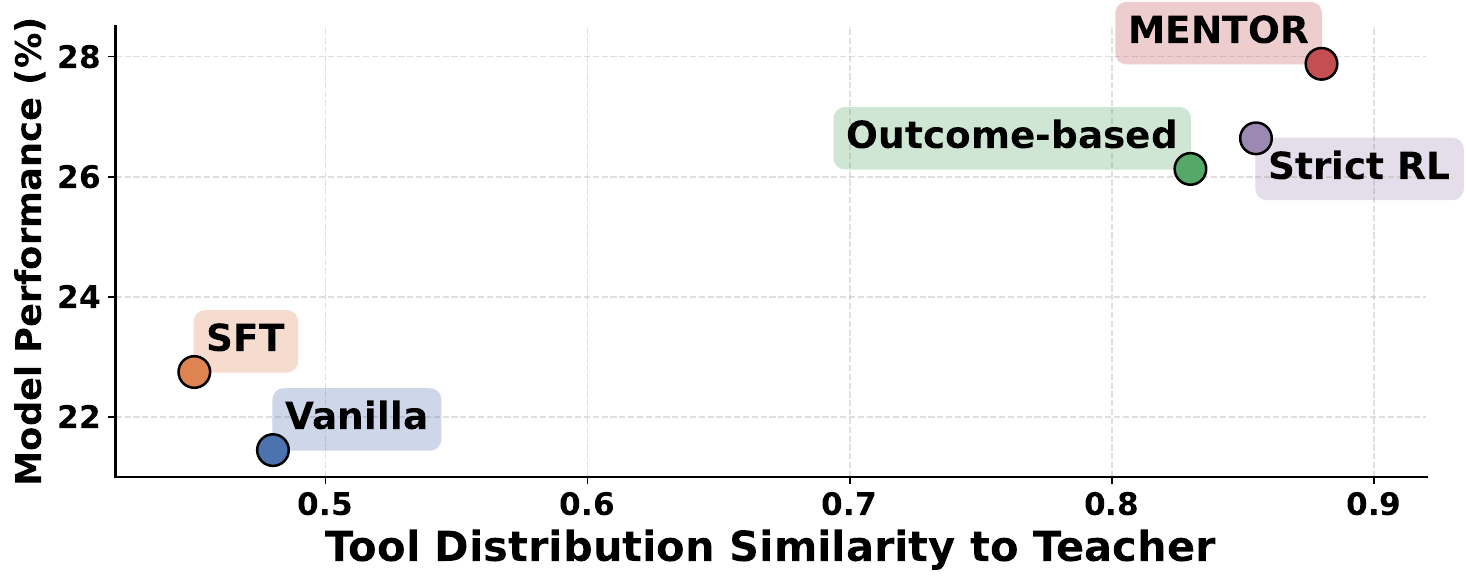}
    \vspace{-0.1in}
    \caption{\label{fig:similarity_and_performance} Correlation between task performance (Math) and alignment score (AS). }
\end{figure}

\begin{figure*}[t]
  \centering
  \includegraphics[width=.9\linewidth]{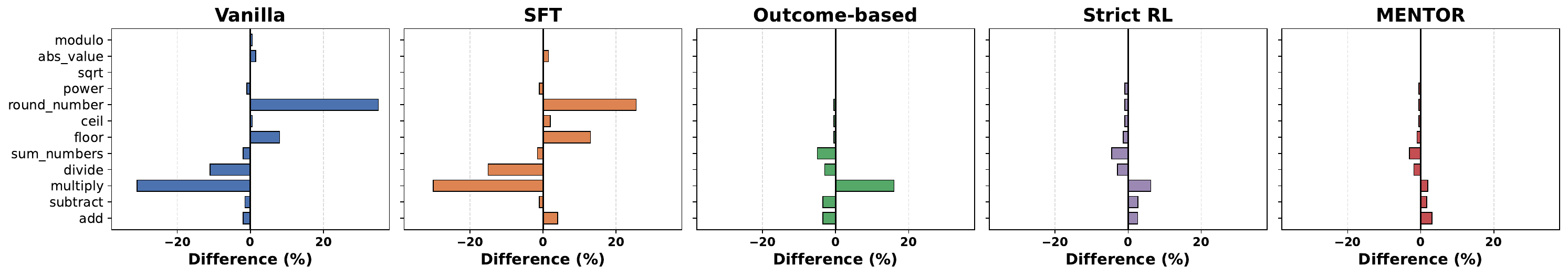}
  \caption{\label{fig:diff_bars} Comparison of tool invocation patterns between student models and the teacher. Each bar represents the percentage point difference in invocation frequency for a specific tool between the student model and the teacher.}
\end{figure*}


\begin{figure}[t]
    \centering
    \begin{subfigure}{0.35\textwidth}
        \centering
        \includegraphics[width=\textwidth]{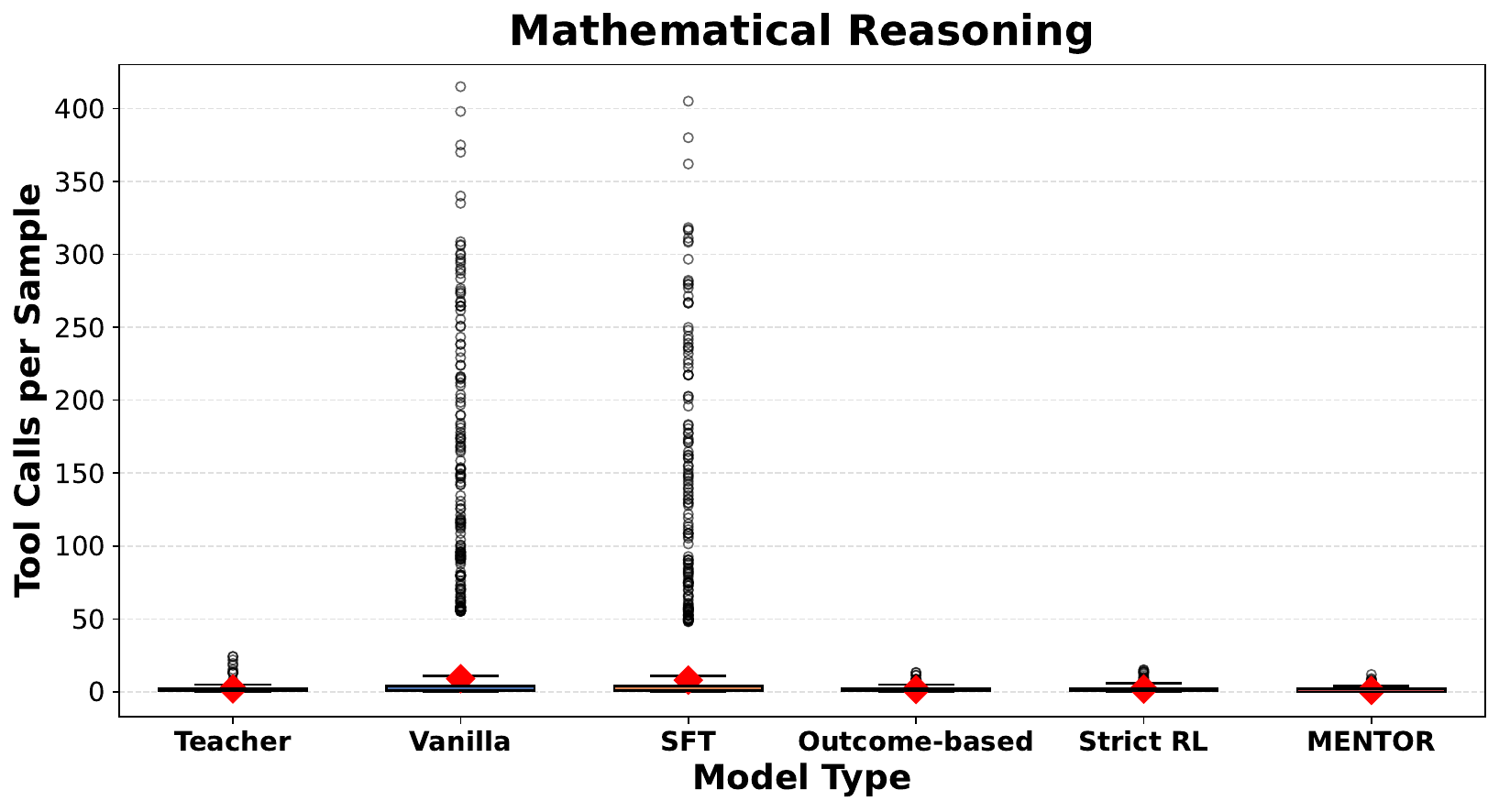}
        \label{fig:tool_count_math}
    \end{subfigure}

    \vspace{-5mm} %
    \begin{subfigure}{0.35\textwidth}
        \centering
        \includegraphics[width=\textwidth]{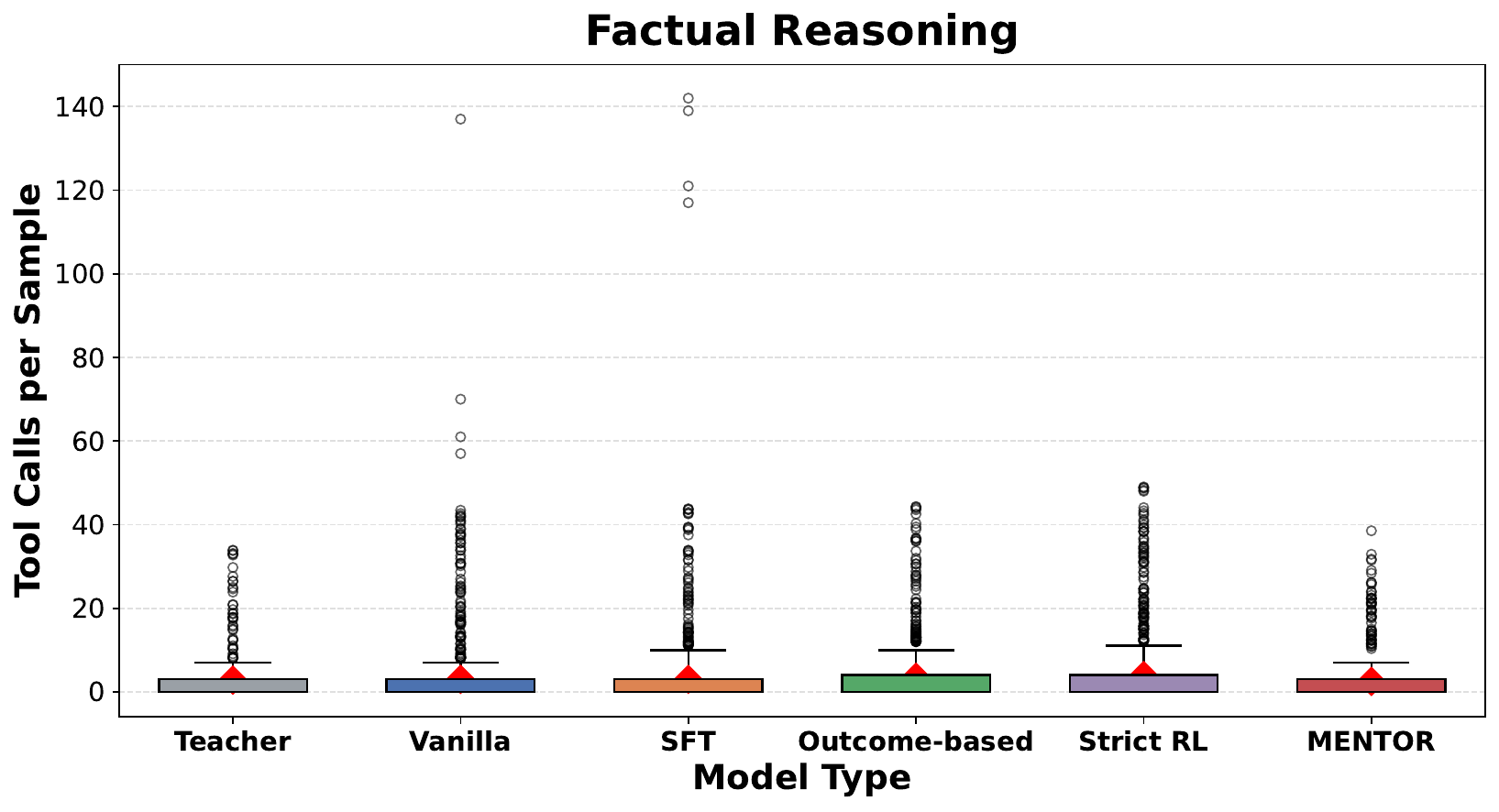}
        \label{fig:tool_count_rag}
    \end{subfigure}
    \vspace{-0.1in} 
    \caption{\label{fig:tool_calling_count_observe} Tool-use efficiency on in-domain and OOD tasks, measured by the distribution of tool calls per sample.}
    \vspace{-0.1in} 
\end{figure}

\paragraph{On-Policy Distillation Achieves General Performance.}
On out-of-domain (OOD) benchmarks, the SFT baseline shows minimal improvement, reflecting the limitations of learning from fixed trajectory distributions. While the Strict RL baseline yields notable gains over SFT, its strict sequential constraints can limit adaptability in unseen tool-use settings. By contrast, \ours{} consistently achieves the highest OOD performance across model scales. This indicates that guiding the model with structural flexibility offers a more reliable approach to generalization than enforcing rigid sequence replication through supervised learning or strict format rewards. For detailed analysis, the following subsections focus on the performance of the Qwen2.5-7B model.

\subsection{Contribution of Flexible Alignment}
\paragraph{Impact of Policy Alignment on Performance.}
We analyze how strategic alignment impacts downstream task performance by plotting model performance against the alignment score (AS) in \autoref{fig:similarity_and_performance}. The results show a clear positive correlation between a student's behavioral similarity to the teacher and their downstream task accuracy, with the SFT baseline achieving the lowest alignment and performance. While RL-based methods generally outperform SFT, \ours{} achieves the highest accuracy and exhibits the closest teacher-aligned tool-use pattern. This simultaneous improvement in both downstream performance and distribution alignment suggests that flexible strategic guidance may be more effective for policy distillation than strict sequence replication.

\paragraph{Distributional Alignment of Tool Usage.}
To evaluate the strategic alignment of each model with the teacher, \autoref{fig:diff_bars} compares each model's tool invocation patterns with the teacher's reference policy. The SFT baseline shows significant distributional deviations, largely retaining the vanilla model's pre-existing invocation habits rather than internalizing the teacher's trajectory distribution. This divergence highlights the inherent limitations of static imitation. In contrast, reinforcement learning baselines narrow the alignment gap, though subtle variations remain observable. While Outcome-based RL exhibits skewed peaks in specific tools, the Strict RL baseline yields relatively low deviations but still leaves subtle, unresolved disparities in overall call frequencies. By contrast, \ours{} demonstrates the tightest and most balanced alignment across all tools. This suggests that \ours{} helps the student adopt tool-selection patterns closer to the teacher while still allowing trajectory-level flexibility during generation.

\begin{figure*}[t]
    \centering
    \begin{minipage}{0.49\linewidth}
        \centering
        \includegraphics[width=0.9\textwidth]{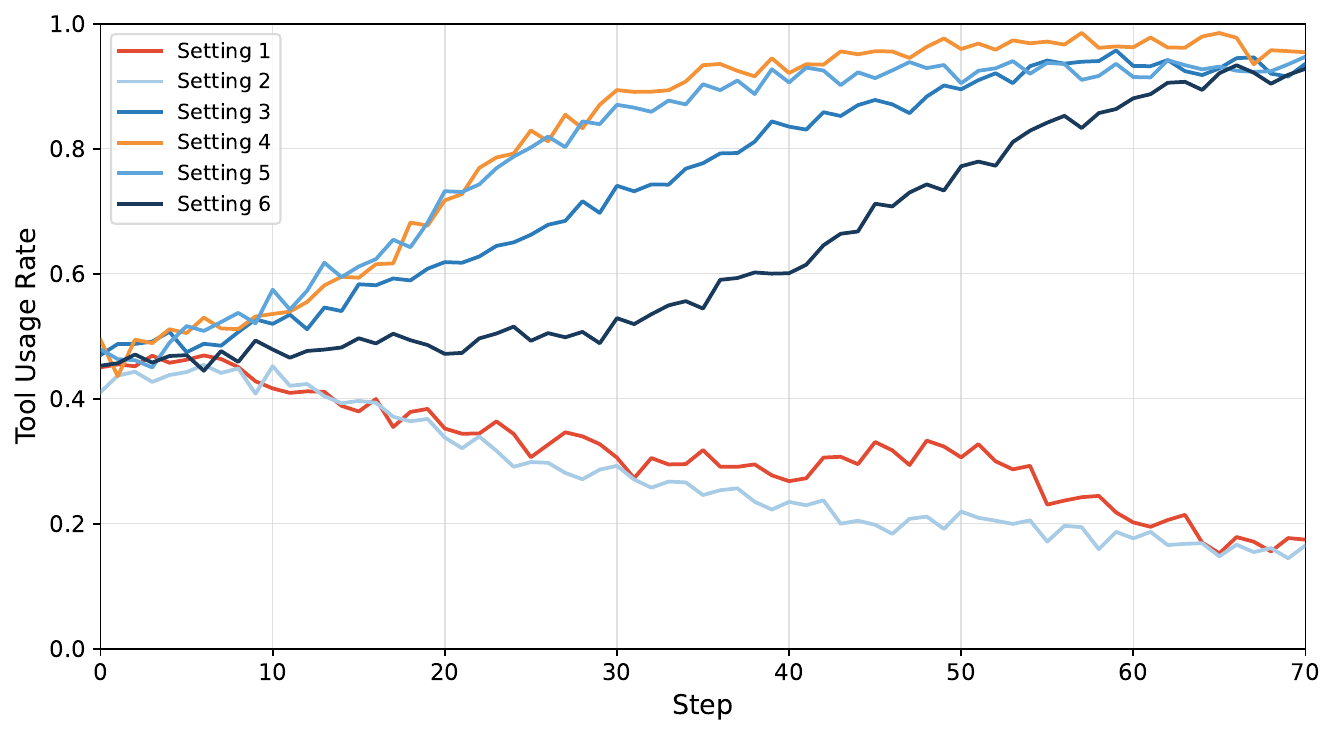}
        \vspace{-0.1in}
        \caption{\label{fig:tool_useage_rate} Tool usage rate over training steps}
    \end{minipage}\hfill
    \begin{minipage}{0.49\linewidth}
        \centering
        \includegraphics[width=0.90\textwidth]{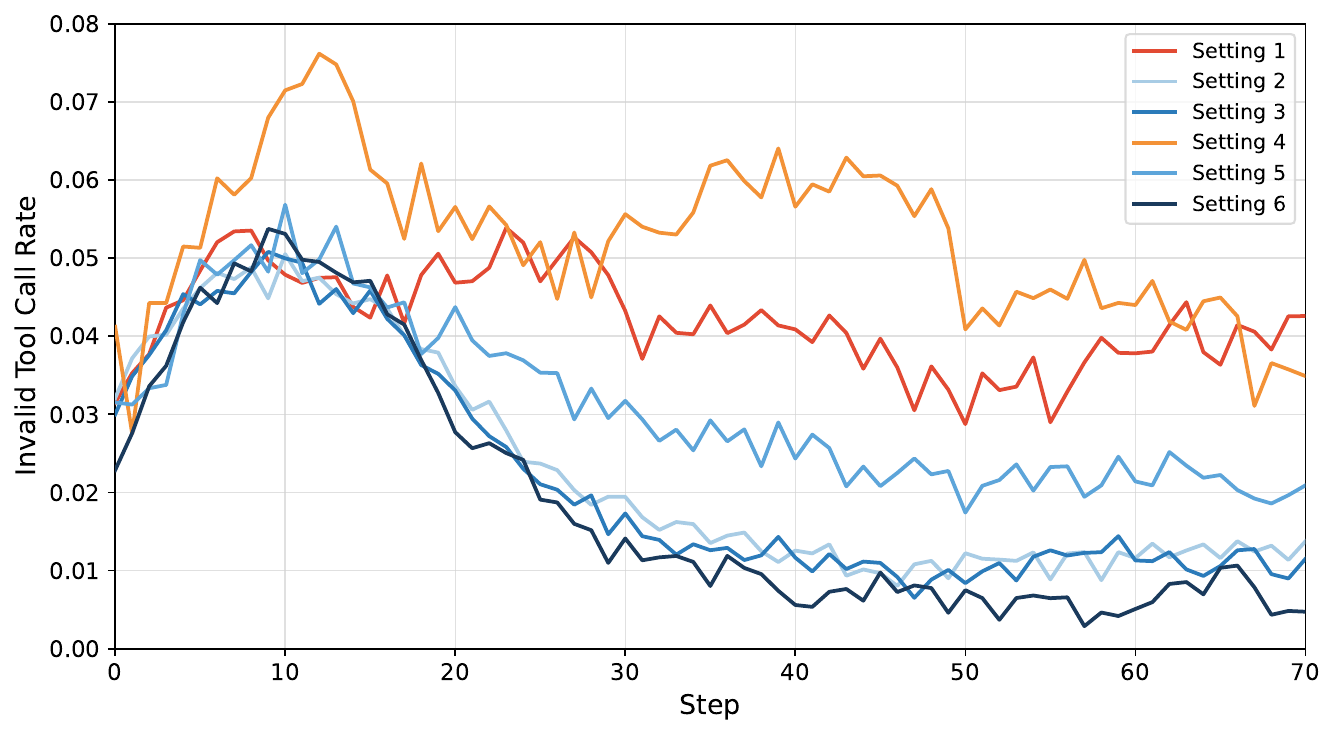}
        \vspace{-0.1in}
        \caption{\label{fig:tool_invalid_rate} Invalid tool call rate over training steps}
    \end{minipage}
\end{figure*}

\paragraph{Behavioral Stability in Tool Usage.}
\autoref{fig:tool_calling_count_observe} shows tool-use efficiency by plotting tool calls per question for in-domain and OOD tasks. The upper outliers and wide spreads in the Vanilla and SFT baselines highlight recurring behavioral errors, such as redundant tool-repetition loops. For in-domain tasks, RL baselines substantially reduce outliers and align with the teacher's concise distribution, indicating that RL optimization is more effective at mitigating invocation errors than SFT. However, in the OOD setting, this trend diverges, with outcome-based and strict RL baselines exhibiting wider spreads and numerous high-frequency outliers. In particular, Strict RL shows more tool repetition and higher variance than outcome-based methods. This divergence suggests that strict sequential constraints may hinder an SLM's response to failures in unseen environments. \ours{}, however, demonstrates stable variance and lower call frequency, suggesting that flexible alignment improves behavioral stability in new tasks.

\begin{table}[t]
\scriptsize
\centering
\setlength{\tabcolsep}{3pt} 
\begin{tabular}{lcccccc}
\toprule
\textbf{Reward Setting} & \textbf{Math} & \textbf{BFCL} & \textbf{RAG} & \textbf{AS} \\
\midrule
(1) $R_\text{c}$ (Outcome-based RL)                                         & 25.94 & 30.71 & 18.22 & 82.51  \\
(2) $R_\text{c}$ + $R_\text{v}$                              & 26.71 & 30.58 & 8.61 & 81.70  \\
(3) $R_\text{a}$ + $R_\text{v}$                    & 24.63 & 26.91 & 14.02 & 78.66  \\
(4) $R_\text{c}$ + $R_\textbf{strict format}$ (Strict RL)                    & 26.37 & 29.92 & 15.61 & 86.28  \\
(5) $R_\text{c}$ + $R_\textbf{strict format}$ + $R_\text{v}$   & 26.82 & 30.47 & 18.19 & 85.87  \\
(6) $R_\text{c}$ + $R_\text{a}$ + $R_\text{v}$ (Ours)             & \bf 27.95 & \bf 31.52 & \bf 21.08 & \textbf{88.63}  \\
\bottomrule
\end{tabular}
\caption{\label{tab:ablation_table} \small Ablation study of the reward
components. Performance is measured by the overall score on the Math, BFCL, and
RAG benchmarks. AS represents the alignment score.}
\vspace{-0.25in}
\end{table}



\subsection{Ablation Study on Reward Components}
\paragraph{Comparative Evaluation of Reward Designs.}
To isolate the contribution of each reward component and compare with prior RL-based TIR frameworks, we evaluate six reward configurations.
The results are shown in \autoref{tab:ablation_table}.
Specifically,  \textbf{Setting (1)} serves as the outcome-based baseline ($R_{\text{c}}$), adapted from ReTool~\cite{feng2025retool}, while \textbf{Setting (2)} adds the tool validation reward ($R_{\text{v}}$) to align with the verification of ARTIST~\cite{singh2025agentic}.
\textbf{Setting (3)} evaluates strategic alignment and tool validity ($R_{\text{a}} + R_{\text{v}}$), omitting the final outcome reward ($R_{\text{c}}$).
\textbf{Setting (4)} incorporates strict format rewards, which aligns with ToolRL~\cite{qian2025toolrl} as our Strict RL baseline.
\textbf{Setting (5)} adds $R_{\text{v}}$ into Setting (4).
Finally, \textbf{Setting (6)} is \ours{}, which employs our flexible, set-based strategic alignment ($R_{\text{a}}$). 
Note that prior reward designs are carefully reimplemented and adapted to fit our experimental framework, ensuring a fair and controlled evaluation.
The results show that our comprehensive reward design (Setting 6) yields the highest overall performance across all tasks and in policy alignment.

\paragraph{Impact of Flexible Rewards on Alignment and Performance.}
The results in \autoref{tab:ablation_table} show a strong positive correlation (Pearson's $r=0.85$) between downstream performance and the alignment score (AS), demonstrating that a flexible reward design effectively optimizes both metrics simultaneously. Crucially, while the Strict RL configurations (Settings 4 and 5) achieve relatively high alignment scores, imposing strict sequential constraints yields diminishing returns on OOD tasks. In contrast, by replacing rigid rules with flexible guidance, \ours{} (Setting 6) achieves the closest policy alignment and the highest generalization accuracy across all tasks. This suggests that a flexible reward structure allows the student to balance strategic internalization with the maintenance of downstream task capabilities.

\paragraph{Teacher-Guided Rewards Foster Robust Tool Adoption.} 
\autoref{fig:tool_useage_rate} illustrates the impact of reward components on the model's tendency to utilize tools. Without the teacher’s guidance (Settings 1 and 2), the sparse reward signal provides insufficient incentive to use tools, causing the student to drift toward a tool-avoidant strategy. Conversely, incorporating teacher alignment (Settings 3-6) actively encourages tool adoption. Crucially, while the strict constraints of the Rigid RL configurations (Settings 4 and 5) drive immediate but rigid usage, \ours{} (Setting 6) demonstrates a steadier learning curve that converges to high usage. This indicates that providing strategic flexibility, despite a slower initial convergence, successfully guides the SLM toward a stable and effective tool-use policy.

\paragraph{Validation Reward Reduces Invocation Errors.}
To evaluate the efficacy of our reward design in mitigating tool invocation errors, \autoref{fig:tool_invalid_rate} shows the invalid tool call rates across ablation configurations during training. The results indicate that the choice of reward components yields distinct learning behaviors. Configurations omitting the tool validation reward ($R_{\text{v}}$), such as Settings 1 and 4, struggle to reduce execution errors and exhibit unstable error rates throughout optimization. Conversely, configurations integrating $R_{\text{v}}$ (Settings 2, 3, 5, and 6) rapidly suppress invalid invocations and converge to near-zero. This suggests that directly penalizing invalid calls with $R_{\text{v}}$ is a highly effective strategy for training a more reliable agent.

\section{Conclusion}
To address the challenges of tool-use distillation arising from the limited capacity of SLMs, we introduce \ours{}. This framework mitigates the reward trade-off between uninformative outcome signals and overly restrictive constraints by combining reinforcement learning with a flexible reward structure. Extensive experiments across multiple reasoning benchmarks demonstrate that \ours{} improves out-of-domain tool-use performance. Our findings suggest that, when training capacity-limited SLMs, providing flexibility to learn within their operational boundaries is more effective than enforcing rigid tool replication. This approach offers a viable path to balancing strategic alignment and downstream task performance.

\section*{Limitations}
While this study offers valuable insights, it is essential to acknowledge that several open challenges remain.

\paragraph{Extending to Other Models.}
Our main experiments focus on the Qwen model series (Qwen2.5 and Qwen3), whose instruction tuning and reasoning-oriented training provide a non-trivial initial ability to produce executable tool-use trajectories. This initial ability is important because our objective is to study reward design for refining and generalizing an existing tool-use policy, rather than to solve tool-call syntax grounding from scratch.

For models that lack reliable initial TIR behavior, directly applying RL can conflate two different problems: learning the basic interface for executable tool calls and optimizing the reward structure for tool-use behavior. In such cases, a short SFT warm-up is a natural initialization step that grounds the model in the tool-use format before RL-based refinement. Importantly, this warm-up is not specific to \ours{}; it should be applied uniformly to all RL-based methods when evaluating models without an initial executable tool-use policy. We therefore view the SFT-then-RL pipeline as a controlled protocol for extending reward-based tool-use distillation to weaker or non-reasoning backbones. A more systematic study of cold-start training strategies remains an important direction for future work.

\paragraph{Scope of Real-World Interactive Environments.}
Our goal in this work is to evaluate reward design for tool-use distillation under controlled executable-tool settings. Accordingly, we evaluate \ours{} in a sandbox environment where tools are invoked through executable code and where tool execution, validity, and final correctness can be measured reliably. This setup is intentionally chosen to isolate the effect of the reward structure itself, rather than to evaluate a full interactive-agent stack.

We emphasize that this scope differs from open-ended real-world interactive environments, such as web browsers, simulators, desktop interfaces, or multi-API systems. Such environments introduce additional environment-level challenges, including interface grounding, noisy observations, long-horizon state tracking, changing tool availability, recovery from failed actions, and safety constraints. These challenges can confound the evaluation of reward design, because performance differences may arise from failures in perception, state tracking, or interface control rather than from the reward signal itself.

Therefore, we leave open-ended interactive environments outside the scope of this study and focus on controlled tool-integrated reasoning settings as a clean testbed for evaluating whether capacity-aware, flexible alignment improves tool-use distillation for SLMs. Extending \ours{} to real-world interactive agents would require combining our reward design with additional mechanisms for environment grounding, action recovery, and long-horizon interaction management, which we view as a separate direction for future work.

\paragraph{Teacher Trajectories as Distillation References.}
\ours{} treats teacher trajectories as distillation references rather than globally optimal tool-use policies. We filter teacher trajectories by retaining only cases where the teacher’s final answer matches the ground truth, which reduces the risk of learning from clearly incorrect demonstrations. Nevertheless, correct teacher trajectories may still be suboptimal in terms of efficiency, ordering, or minimal tool use. This issue reflects the boundary of our study: we focus on reward design for teacher-guided distillation, not on directly discovering optimal tool-use policies from ground-truth supervision or environment-level search. Within this distillation setting, teacher trajectories provide a useful behavioral prior for guiding capacity-limited SLMs, while final-correctness and tool-validation rewards prevent the student from being rewarded for invalid or unsuccessful solutions. Exploring direct optimization toward ground-truth optimal trajectories remains a complementary direction for future work.

\paragraph{Scope of Source-Domain Training.}
We adopt mathematical reasoning as the source domain because it offers one of the most suitable settings for studying tool-use distillation: it supports diverse multi-step TIR paths, provides executable tool traces, and enables reliable evaluation of both final correctness and tool validity. Domains with similarly rich and verifiable TIR trajectories could also be used to train MENTOR-style teacher-guided rewards. In our experimental setting, mathematical reasoning was the most suitable training source among the considered benchmarks because it naturally provides dense multi-step TIR trajectories, executable traces, and reliable correctness signals. We leave systematic validation across broader source domains as future work.

\paragraph{Computational Constraints on Reward Weight Search.}
A full grid search over \(w_c\), \(w_a\), and \(w_v\) is computationally expensive because each configuration requires complete RL training with multiple rollouts. We therefore select the final configuration using development-set diagnostics and report sensitivity and ablation analyses in Appendix~\ref{appen:weights}. More systematic reward-weight optimization remains a direction for future work.

\section*{Ethical Statements}
This work contributes to the development of efficient and capable artificial intelligence. By successfully distilling the complex tool-use capabilities of large language models (LLMs) into smaller, more efficient small language models (SLMs), \ours{} accelerates the creation of functional, on-device AI. This capability enables local deployment, reducing reliance on expensive cloud infrastructure and improving user privacy for agents that perform complex tasks, such as mathematical reasoning and information retrieval from external sources (including the web).

However, the enhanced strategic competence and tool-augmented abilities conferred by \ours{} also introduce potential risks. Since our distilled agents are capable of autonomous reasoning, web retrieval, and code execution, they could be susceptible to misuse. Potential malicious behaviors include the automated generation of harmful scripts, the execution of unauthorized actions, or the spread of misinformation via tool-based retrieval. To ensure responsible deployment, the integration of robust safeguards is essential. We emphasize that addressing these ethical and safety concerns is an important direction for future research and responsible development in the field of small language agents.

\section*{Acknowledgements}
This research was supported by the Institute of Information \& communications Technology Planning \& Evaluation (IITP) grant, funded by the Korea government (MSIT) (No. RS-2026-25525363, AEGIS: Agentic Experts for Generative-AI Inspection Solution), and by the National Research Foundation of Korea (NRF) grant funded by the Korea government (MSIT) (No. RS-2026-25491073, A Reinforcement Learning-Based General-Purpose GUI Agent for Aligning Human Decision-Making Trajectories). We also gratefully acknowledge LG CNS for supporting this collaborative research. This work utilized GPU resources from the ``Advanced GPU Utilization Support Program'' funded by MSIT, Republic of Korea (awarded to KyungTae Lim).

We used AI assistants, including ChatGPT\footnote{\url{https://chatgpt.com/}}, Gemini\footnote{\url{https://gemini.google.com}} and Grammarly\footnote{\url{https://app.grammarly.com/}}, to support the writing and coding processes.


\bibliography{custom}
\clearpage

\appendix

\section{\label{appen:example_case}Examples of Tool Failures on SLMs}
\subsection{\label{appen:alignment}Examples of Discrepancy in Tool Selection}

We provide a qualitative example to illustrate the tool-use behavior discussed in Section~3. The example compares the teacher LLM and the student SLM on the same mathematical reasoning problem. It is intended to complement the aggregate results in \autoref{fig:observation}, rather than serve as standalone evidence.

\autoref{tab:llm_example} shows that the teacher model follows a concise tool-use trajectory. It computes \(g(16)=\sqrt{16}=4\), then \(g(4)=\sqrt{4}=2\), and finally uses the subtract tool to obtain \(t(g(16))=3-2=1\). The tool calls directly correspond to the necessary computational steps, and the model commits to the final answer after obtaining the relevant intermediate values.

\autoref{tab:slm_example} shows a different pattern. The student model initially reaches the correct value using the same core tools, but it does not terminate the trajectory after the necessary computation is complete. Instead, it issues redundant verification calls that are themselves erroneous---swapping the arguments of one subtraction and dropping a sign in an addition---which produce conflicting values. Rather than recognizing that its original computation was already correct, the student trusts the latest, incorrect result and commits a wrong final answer. This suggests that the student struggles not only with selecting tools, but also with judging when the accumulated tool evidence is sufficient, so that unnecessary further tool use can actively corrupt an already-correct solution.

This example illustrates that the performance gap is not merely a matter of whether the student invokes tools. The student can call relevant tools and even derive the correct intermediate value, but may still struggle to decide when to stop, so that additional unnecessary tool calls introduce new errors rather than confirming the answer. These behaviors support our motivation for a reward design that jointly encourages correct task completion, teacher-guided tool selection, and valid tool execution.

\begin{figure}[t]
    \centering
    \includegraphics[width=\linewidth]{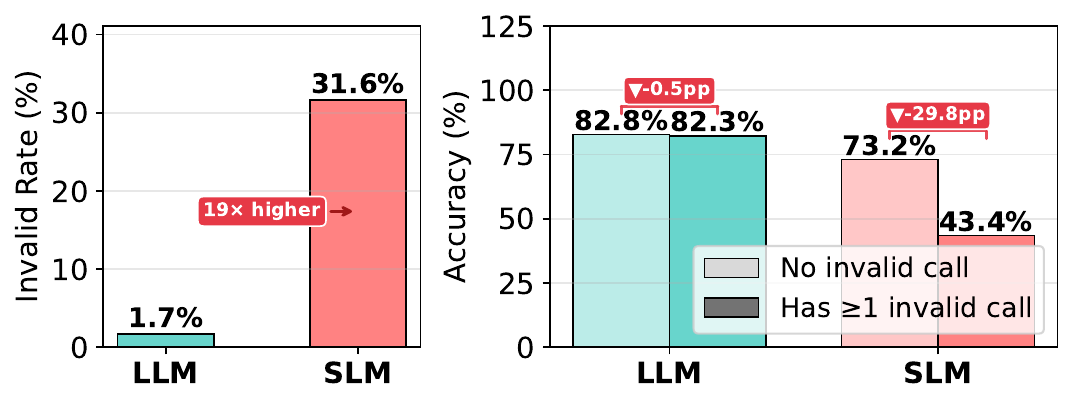}
    \caption{\label{fig:appen_obs_invalid}Invalid call rate (left) and its impact on accuracy (right).}
\end{figure}

\subsection{The Critical Impact of Invalid Tool Calls on SLM Performance.}
As illustrated in \autoref{fig:appen_obs_invalid}, small language models exhibit an extreme vulnerability to tool execution failures compared to larger models. On the left, the baseline SLM generates invalid tool calls at a rate of 31.6\%, which is nearly 19 times higher than that of the teacher LLM (1.7\%). More critically, as shown on the right, there is a stark contrast in error recovery capabilities between the two models. While the teacher LLM maintains its performance with a negligible accuracy drop of only 0.5 percentage points when encountering execution errors, a single invalid tool call causes the SLM's accuracy to plunge from 73.2\% to 43.4\%—a severe drop of 29.8 percentage points.

This catastrophic performance degradation indicates that the SLM lacks the intrinsic capability to self-correct or replan once its initial reasoning path is disrupted by a malformed tool call. Instead of diagnosing the execution error and generating a corrective alternative path, the smaller model enters a cascade of failures. Consequently, these findings underscore that bridging the performance gap in tool-integrated reasoning requires more than simply cloning optimal teacher trajectories through sequential imitation. To build true execution resilience, the SLM must internalize the underlying strategic planning logic of the teacher, enabling it to sustain coherent reasoning even when individual tool steps fail.

%

\begin{table*}[t]
    \centering
    \small
    \begin{tcolorbox}[title={LLM (Qwen3-235B) Trajectory --- Sample~\#149, MATH Algebra L3},
                      width=0.95\textwidth, colback=white, colframe=black!60, sharp corners=south]
        \textbf{QUESTION:}
        \begin{lstlisting}[basicstyle=\small\ttfamily, breaklines=true]
If t(x) = 3 - g(x) and g(x) = sqrt(x), then what is t(g(16))?

(If you have got the answer, enclose it within \boxed{} with latex format.)

Ground truth: 1
        \end{lstlisting}
        \vspace{0.6em}

        \textbf{ASSISTANT (reasoning):}
        \begin{lstlisting}[basicstyle=\small\ttfamily, breaklines=true]
<think>
... t(g(16)) = 3 - g(g(16)). First, g(16) = sqrt(16) = 4.
Then g(4) = sqrt(4) = 2. So t(g(16)) = 3 - 2 = 1.

Looking at the tools, sqrt and subtract are available, so:
  1. Call sqrt(16)  -> 4
  2. Call sqrt(4)   -> 2
  3. Call subtract(3, 2) -> 1

So the tools needed are sqrt (twice) and subtract (once).
</think>
        \end{lstlisting}

        \textbf{ASSISTANT (tool calls, issued in parallel in a single turn):}
        \begin{lstlisting}[basicstyle=\small\ttfamily, breaklines=true]
<tool_call>{"name": "sqrt",     "arguments": {"number": 16}}</tool_call>
<tool_call>{"name": "sqrt",     "arguments": {"number": 4}}</tool_call>
<tool_call>{"name": "subtract", "arguments": {"minuend": 3, "subtrahend": 2}}</tool_call>
        \end{lstlisting}

        \textbf{TOOL RESPONSES:}
        \begin{lstlisting}[basicstyle=\small\ttfamily, breaklines=true]
<tool_response>{"name": "sqrt", "arguments": {"number": 16}}
  result: 4.0
</tool_response>
<tool_response>{"name": "sqrt", "arguments": {"number": 4}}
  result: 2.0
</tool_response>
<tool_response>{"name": "subtract", "arguments": {"minuend": 3, "subtrahend": 2}}
  result: 1
</tool_response>
        \end{lstlisting}

        \textbf{ASSISTANT (final answer):}
        \begin{lstlisting}[basicstyle=\small\ttfamily, breaklines=true]
<think>
Everything checks out. The answer should be 1.
</think>

\boxed{1}
        \end{lstlisting}
    \end{tcolorbox}
    \caption{\label{tab:llm_example}LLM trajectory for sample~\#149 (MATH Algebra L3). The teacher (Qwen3-235B) plans the full computation in a single reasoning step, issues the three required tool calls (\texttt{sqrt}, \texttt{sqrt}, \texttt{subtract}) in parallel within one turn, and commits the boxed answer as soon as the results return. This concise, decisive pattern contrasts with the SLM in Table~\ref{tab:slm_example}, which reaches the same value but then derails into redundant, erroneous verification calls and commits a wrong final answer.}
\end{table*}

\clearpage
%

\begin{table*}[t]
    \centering
    \small
    \begin{tcolorbox}[title={SLM (Qwen3-8B) Trajectory --- Sample~\#149, MATH Algebra L3},
                      width=0.95\textwidth, colback=white, colframe=black!60, sharp corners=south]
        \textbf{QUESTION:}
        \begin{lstlisting}[basicstyle=\small\ttfamily, breaklines=true]
If t(x) = 3 - g(x) and g(x) = sqrt(x), then what is t(g(16))?
Ground truth: 1
        \end{lstlisting}
        \vspace{0.4em}

        \textbf{Initial solve (turns 1--3):} three tool calls correctly yield $1.0$.
        \begin{lstlisting}[basicstyle=\small\ttfamily, breaklines=true]
<think> g(16) = sqrt(16). Let me check. </think>
<tool_call>{"name": "sqrt", "arguments": {"number": 16}}</tool_call>           -> 4.0

<think> Now g(4.0) = sqrt(4.0) = 2.0. </think>
<tool_call>{"name": "sqrt", "arguments": {"number": 4.0}}</tool_call>          -> 2.0

<think> So t(4.0) = 3 - 2.0 = 1.0. </think>
<tool_call>{"name": "subtract", "arguments": {"minuend": 3, "subtrahend": 2.0}}</tool_call>  -> 1.0
        \end{lstlisting}

        \textit{\textcolor{red!70!black}{The correct answer~(1.0) is now available. A confident agent would commit \texttt{\textbackslash boxed\{1\}} here. The SLM does not.}}
        \vspace{0.4em}

        \textbf{Verify-loop (turns 4--5):} two redundant calls, issued out of self-doubt, are themselves erroneous.
        \begin{lstlisting}[basicstyle=\small\ttfamily, breaklines=true]
<think> Let me double-check the subtraction the other way. </think>
<tool_call>{"name": "subtract", "arguments": {"minuend": 2.0, "subtrahend": 3}}</tool_call>  -> -1.0  (arguments swapped: WRONG)

<think> That gives -1.0, not 1.0. Let me try once more with add. </think>
<tool_call>{"name": "add", "arguments": {"firstNumber": 3, "secondNumber": 2.0}}</tool_call>  -> 5.0  (sign error, should be -2.0: WRONG)
        \end{lstlisting}
        \vspace{0.4em}

        \textbf{Final turn (turn 6) --- commits the wrong value:}
        \begin{lstlisting}[basicstyle=\small\ttfamily, breaklines=true]
<think> My checks disagree (1.0, -1.0, 5.0); the latest gave 5.0. </think>

\boxed{5}
        \end{lstlisting}
        \vspace{0.4em}
        \textbf{Final answer submitted: 5} --- incorrect (ground truth is~1).
        The model abandoned its already-correct result~(1.0) and trusted a
        later, erroneous tool call.
    \end{tcolorbox}
    \caption{\label{tab:slm_example}SLM trajectory for sample~\#149 (condensed; \texttt{<think>} blocks are shortened to the model's key sentence per turn). The student derives the correct value~(1.0) at turn~3 but does not commit. Instead, driven by self-doubt rather than computational need, it issues two redundant verification calls that are themselves erroneous --- swapping the arguments of \texttt{subtract} (yielding $-1.0$) and dropping a sign in \texttt{add} (yielding $5.0$). Faced with the conflicting results, it abandons its already-correct answer and commits the latest, wrong value (\texttt{\textbackslash boxed\{5\}}), so the sample is marked incorrect. Total: 6 turns, 5 tool calls (2 redundant and erroneous). This verify-loop pattern contributes additional \texttt{add}/\texttt{subtract} calls to the SLM's tool distribution shown in Figure~\ref{fig:observation}, and illustrates how unnecessary tool use can actively corrupt an already-correct solution.}
\end{table*}

\clearpage

\begin{figure*}[t]
\begin{equation}
\label{grpo_equation}
\small
    \begin{array}{l}
    \mathcal{J}_{GRPO}(\theta)=\mathbb{E}_{\substack{x_i \sim \mathcal{D},\text{ } \{O^{(s)}_j\}_{j=1}^G \sim \pi_{old}(\cdot|x_i)}}\Biggl[\frac{1}{G} \sum_{j=1}^{G} \left({\sum_{k=1}^{|\tau_j^{(s)}|} \mathbb{I}(\tau_{j,k})}\right)^{-1}\\                 
  \sum_{k=1}^{|\tau_j^{(s)}|} \min\left(\frac{\pi_{\theta}(\tau_{j,k} | \tau_{j,<k}, x_i)}{\pi_{old}(\tau_{j,k} | \tau_{j,<k}, x_i)} \hat{A}_{j,k}, clip(\frac{\pi_{\theta}(\tau_{j,k} | \tau_{j,<k}, x_i)}{\pi_{old}(\tau_{j,k} | \tau_{j,<k}, x_i)},1-\epsilon,1+\epsilon)\hat{A}_{j,k}\right)\cdot \mathbb{I}(\tau_{j,k})
  - \beta \mathbb{D}_{KL}[\pi_{\theta}|| \pi_{ref}]\Biggr]   
    \end{array}
\end{equation}
\vspace{-0.2in}
\end{figure*}

\section{Details on \ours{}}
\subsection{\label{appen:method_detail}Adapting GRPO for Reasoning Distillation}
We adapt the GRPO algorithm to conduct on-policy distillation by treating the teacher's trajectory as an alignment reference rather than a token-level supervision target. This reward-driven setup trains the student to internalize the teacher's strategic tool-use policy, rather than merely imitating a fixed sequence of actions.
This process is detailed in Algorithm~\ref{algor:grpo}.

\paragraph{Generating Rollouts}
For each question $x_i$ in our training set, we perform two simultaneous generation steps. First, we retrieve the corresponding reference trajectory, $(\tau^{(t)},\hat{y}^{(t)})$, which was generated by the teacher model as described in Section~\ref{sec:method_overview}.
Then, we use the current student model ($\pi_{old}$) to generate a group of $G$ rollouts. This step is crucial as it allows the student to explore diverse reasoning paths and tool-use strategies for the same problem, providing the varied data needed for the GRPO comparison.

\paragraph{Student Policy Optimization}
The core idea of our proposed method is to update a policy by aligning a high-quality reference against a group of sampled candidates.
We optimize the tool-use policy of the student model based on the reference trajectory generated by the teacher. We optimize the student model by maximizing the objective function as shown in \autoref{grpo_equation}, where $\epsilon$ and $\beta$ are hyperparameters for clipping and KL regularization, respectively.
The advantage, $\hat{A}_{j,k}$, is computed from the relative rewards within the sample group. The reference policy $\pi_{ref}$ used for KL regularization is the initial student model before RL training.
$\mathbb{I}(\tau_{j,k})$ is an indicator function used for loss masking, which equals 1 if $\tau_{j,k}$ is an LLM-generated token, and 0 otherwise.

\begin{algorithm}[t]
\scriptsize
\caption{\label{algor:grpo}Training with GRPO}
\label{alg:artist-grpo}
\begin{algorithmic}[1]
\Require Student model $\pi_\theta$, old student model $\pi_{\text{old}}$, Teacher model $\pi_\text{teacher}$, task dataset $\mathcal{D}$, group size $G$, indicator function $\mathbb{I}$
\For{each training iteration}
  \For{each question $x_i$}
    \State Generate reference trajectory $O^{(t)}$ from $\pi_\text{teacher}$
    \State Sample $G$ rollouts $\{O^{(s)}_1,\dots,O^{(s)}_G\}$ from $\pi_\text{old}$
    \For{each rollout $O^{(s)}_j$}
        \State Compute outcome rewards $R(O^{(s)}_j, O^{(t)}))$
    \EndFor
    \State Compute groupwise advantages $\hat{A}_{j,k}$ for all $O^{(s)}_j$ 
    \State Apply indicator $\mathbb{I}$ to mask tool output tokens
    \State Compute GRPO loss $\mathcal{L}_{\text{GRPO}}$ and update $\pi_\text{student}$
  \EndFor
\EndFor
\end{algorithmic}
\end{algorithm}

\subsection{\label{appen:weights}Reward Weight Optimization}
The composite reward in \autoref{eq:composite-reward} is scaled by three hyperparameters, $w_c$, $w_a$, and $w_v$, which balance the relative contributions of correctness, strategic alignment, and tool validation signals, respectively. Because each reward configuration requires full RL training with multiple rollouts, exhaustive weight search is computationally expensive. We therefore adopt an empirical tuning protocol to determine the hyperparameters under these resource limitations.

\paragraph{Tuning Protocol.}
We monitor the values of each reward component throughout the training process, alongside the overall learning curve. We adjust the weights under two specific conditions: (i) when a single reward component becomes overly dominant and suppresses the signals from the other components, or (ii) when training becomes unstable, as indicated by fluctuating tool usage or a sudden increase in invalid calls. The goal of this tuning is not to maximize any single metric, but to maintain a balanced reward system where all three signals jointly guide the model's policy.

\paragraph{Sensitivity Analysis.}
To validate our final configuration of $(w_c, w_a, w_v) = (0.7, 0.3, 0.1)$, we conduct a controlled sensitivity sweep by varying each weight independently while holding the other two fixed, alongside a uniform-weight baseline for comparison. The results are presented in \autoref{tab:reward-weight-ablation}, where Math, BFCL, and RAG represent the overall task scores, and AS denotes the alignment score.

\begin{table}[h]
\centering
\small
\setlength{\tabcolsep}{4pt}
\renewcommand{\arraystretch}{1.15}
\begin{tabular}{lcccc}
\toprule
$(w_c, w_a, w_v)$ & Math & BFCL & RAG & AS \\
\midrule
\multicolumn{5}{l}{\emph{Reference (ours)}} \\
$(0.7,\ 0.3,\ 0.1)$            & \textbf{27.95} & \textbf{31.52} & \textbf{21.08} & 88.63 \\
\midrule
\multicolumn{5}{l}{\emph{Varying $w_c$ (correctness)}} \\
$(1.0,\ 0.3,\ 0.1)$            & 27.69 & 30.94 & 19.92 & 86.71 \\
$(0.5,\ 0.3,\ 0.1)$            & 26.84 & 30.75 & 20.22 & 88.21 \\
\midrule
\multicolumn{5}{l}{\emph{Varying $w_a$ (teacher alignment)}} \\
$(0.7,\ 0.1,\ 0.1)$            & 27.32 & 30.51 & 20.04 & 86.43 \\
$(0.7,\ 0.5,\ 0.1)$            & 27.31 & 31.02 & 20.54 & \textbf{88.95} \\
\midrule
\multicolumn{5}{l}{\emph{Varying $w_v$ (tool validation)}} \\
$(0.7,\ 0.3,\ 0.0)$            & 26.97 & 30.18 & 18.62 & 87.02 \\
$(0.7,\ 0.3,\ 0.3)$            & 27.51 & 31.08 & 20.86 & 88.34 \\
\midrule
\multicolumn{5}{l}{\emph{Uniform reference}} \\
$(0.33,\ 0.33,\ 0.33)$         & 27.06 & 30.74 & 20.41 & 88.05 \\
\bottomrule
\end{tabular}
\caption{Sensitivity of \ours{} to the reward weights $(w_c, w_a, w_v)$. Each
group perturbs a single axis while holding the others fixed; the final row
applies uniform weights as an additional reference. Math, BFCL, and RAG
report the overall task scores; AS is the alignment score
(Appendix~\ref{appen:metric}).Each configuration is independently trained once and evaluated over three runs,
whereas \autoref{tab:main_results} reports means over ten runs.}
\label{tab:reward-weight-ablation}
\end{table}

\paragraph{Observations.}
The sensitivity analysis reveals three key trade-offs that justify our final hyperparameter selection. 
First, along the correctness axis ($w_c$), increasing $w_c$ to $1.0$ provides no further benefits for Math ($-0.26$) and significantly degrades both OOD performance ($-0.58$ on BFCL, $-1.16$ on RAG) and policy alignment ($-1.92$ AS). This indicates that correctness optimization is already saturated at $w_c = 0.7$, and further amplification only weakens the strategic alignment signal. Conversely, reducing $w_c$ to $0.5$ causes the largest in-domain performance drop ($-1.11$ on Math) while leaving the alignment score relatively unchanged ($-0.42$ AS). This confirms that task correctness is the primary driver of task success.

Second, along the teacher alignment axis ($w_a$), decreasing $w_a$ to $0.1$ results in the most severe alignment degradation in the table ($-2.20$ AS), alongside a notable decrease on BFCL ($-1.01$). Pushing $w_a$ to $0.5$ improves the alignment score by $+0.32$ AS but trades away task accuracy across all benchmarks. This directly mirrors the well-known alignment-versus-utility trade-off, confirming that our moderate setting of $w_a = 0.3$ serves as a balanced joint optimum. 

Third, along the tool validation axis ($w_v$), setting $w_v$ to $0.0$ produces the lowest RAG score in the table ($-2.46$) and a clear drop on BFCL ($-1.34$). This drop is consistent with the higher rate of execution failures observed when validation rewards are absent. However, increasing $w_v$ to $0.3$ yields no additional task performance gains, indicating that a small but non-zero penalty is sufficient to suppress malformed tool calls without making the model's policy overly conservative.

Finally, the uniform-weight baseline $(0.33, 0.33, 0.33)$ achieves a comparable alignment score ($-0.58$ AS) but underperforms the reference configuration on every task metric. This reinforces that our final configuration is not dominated by a single component, but rather emerges from the joint balance of all three signals.

We also note that the main ablation study in \autoref{tab:ablation_table} implicitly evaluates extreme weight configurations: Settings (1)–(2) correspond to $w_a = 0$ and Setting (3) corresponds to $w_c = 0$. Because both extremes underperform \ours{} (Setting 6), these results reinforce the conclusion that all three reward components play complementary, non-redundant roles in policy distillation.

\subsection{\label{appen:set_base}Comparison of the Set-based Approach with Alternative Methods}

\paragraph{Motivation for Flexible Teacher Alignment.}
The teacher-alignment reward is designed to provide intermediate guidance between the two undesirable extremes observed in Section~3: sparse outcome-only rewards, which provide insufficient incentive for tool use, and strict trajectory matching, which can over-constrain capacity-limited SLMs. Our goal is therefore not to force the student to reproduce the teacher's trajectory exactly, but to provide a flexible alignment signal that encourages the student to use the teacher-selected tools while allowing the execution order and trajectory structure to adapt to the student's capacity. This motivates a reward that is permissive to order-level variations in how the same teacher-selected tools are executed, while remaining informative enough to distinguish trajectories that preserve the teacher's high-level tool-selection behavior from those that drift toward tool avoidance or unstable tool invocation.

\paragraph{Alternative Flexible Alignment Rewards.}
The set-based reward \(R_a\) used in \ours{} is one instantiation of a broader family of flexible teacher-alignment rewards. To examine whether the benefit comes from the general principle of flexible alignment or from the specific set-matching formulation, we compare five variants that differ in order sensitivity, multiplicity, and reward granularity. Let \(\tau^{(s)}\) and \(\tau^{(t)}\) denote the student and teacher tool sequences, \(S(\cdot)\) extract the set of distinct tools, and \(M(\cdot)\) extract the multiset of tools with call counts.

\begin{itemize}[leftmargin=1.2em,itemsep=2pt,topsep=2pt]
\item \textbf{Set matching (\ours{}).} This unordered, binary, presence-based reward gives credit when the student uses the same set of tools as the teacher, regardless of execution order or call counts:
\[
R_a^{\mathrm{set}} = \mathbb{1}\!\left[\,S(\tau^{(s)}) = S(\tau^{(t)})\,\right].
\]

\item \textbf{Ordered matching.} This variant adds an order constraint by comparing the student and teacher tool sequences position-wise while ignoring tool arguments:
\[
R_a^{\mathrm{ord}} = \mathbb{1}\!\left[\,\tau^{(s)} = \tau^{(t)}\,\right].
\]
It tests whether preserving the teacher's tool order is beneficial beyond matching tool choice.

\item \textbf{Multiset matching.} This variant is order-agnostic but preserves call counts:
\[
R_a^{\mathrm{multi}} = \mathbb{1}\!\left[\,M(\tau^{(s)}) = M(\tau^{(t)})\,\right].
\]
It tests whether matching the number of invocations for each tool improves alignment compared with presence-only matching.

\item \textbf{F1 (graded set).} This variant replaces binary set equality with a graded overlap score:
\[
R_a^{\mathrm{F1}} =
\frac{2\,|S(\tau^{(s)}) \cap S(\tau^{(t)})|}
{|S(\tau^{(s)})| + |S(\tau^{(t)})|}.
\]
It provides partial credit when the student recovers only part of the teacher's tool set.

\item \textbf{Recall-only.} This variant rewards coverage of the teacher's tools without penalizing additional tools:
\[
R_a^{\mathrm{rec}} =
\frac{|S(\tau^{(s)}) \cap S(\tau^{(t)})|}
{|S(\tau^{(t)})|}.
\]
It tests whether alignment should focus only on missing teacher tools or also discourage unnecessary extra tools.
\end{itemize}

All five variants are combined with the same correctness reward \(R_c\) and tool-validation reward \(R_v\), and are trained under the same GRPO setup.

\begin{figure}[t]
    \centering
    \begin{subfigure}[t]{\columnwidth}
        \centering
        \includegraphics[width=0.8\textwidth]{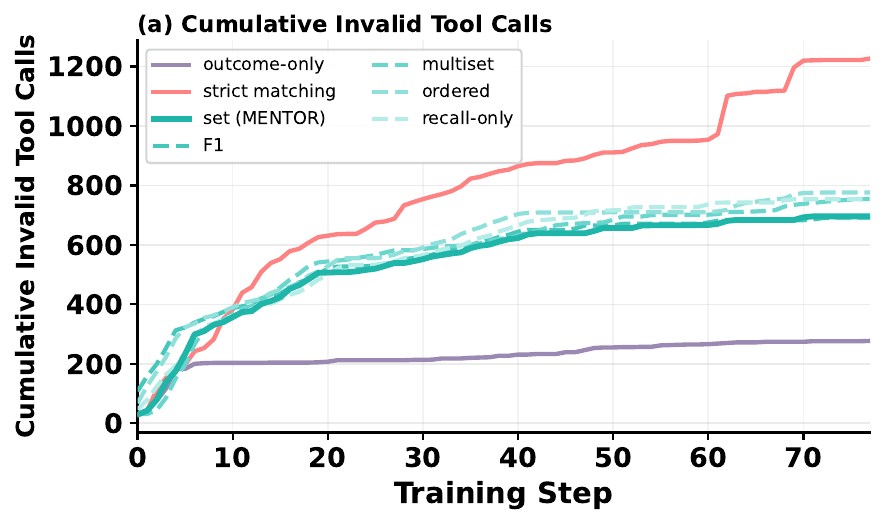}
        \vspace{-0.2in}
        \label{fig:b3_flex_invalid}
    \end{subfigure}

    \vspace{1mm}
    \begin{subfigure}[t]{\columnwidth}
        \centering
        \includegraphics[width=0.8\textwidth]{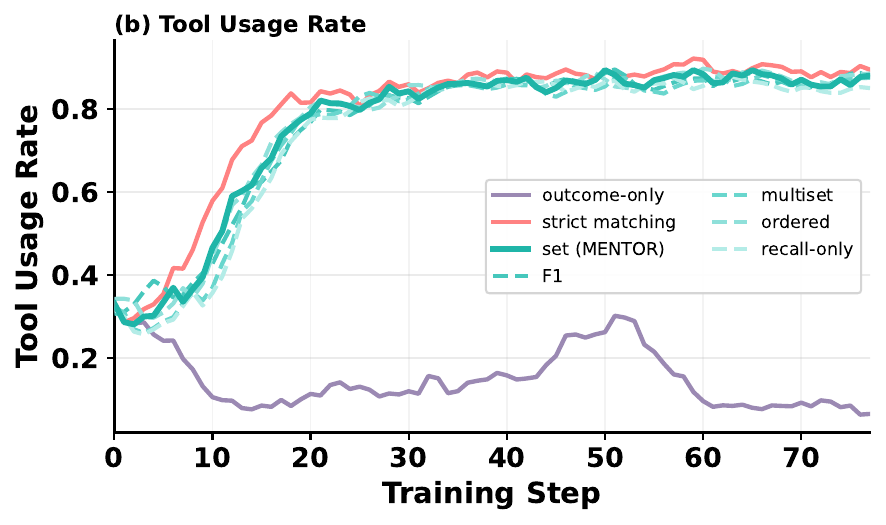}
        \vspace{-0.2in}
        \label{fig:b3_flex_usage}
    \end{subfigure}
    \vspace{-0.3in}
    \caption{\label{fig:b3_flexible_variants}\small Training dynamics of the five flexible-alignment variants, with the outcome-only and strict-matching references from Figure~\ref{fig:observation_learning}. (a) Cumulative invalid tool calls and (b) tool-usage rate over training steps. The five flexible variants cluster tightly together---all suppressing invalid calls relative to strict matching while sustaining high tool usage, unlike the tool-avoidant outcome-only run---indicating that the learning dynamics are governed by the flexible-alignment principle rather than the specific variant.}
\end{figure}

\paragraph{Training Dynamics Across Flexible Variants.}
Figure~\ref{fig:b3_flexible_variants} compares the training dynamics of the five flexible alignment variants with the outcome-only and strict-matching references from Figure~\ref{fig:observation_learning}. The flexible variants show broadly similar behavior: they sustain a high tool-usage rate, unlike the outcome-only baseline, while keeping invalid tool calls substantially lower than strict matching. This pattern suggests that the main trade-off identified in Section~3 can be addressed by a family of flexible teacher-alignment rewards, rather than by set matching alone. In other words, the key mechanism is to provide teacher-guided process feedback that preserves the teacher-selected tool set without enforcing its exact execution order.

\begin{table}[t]
\centering
\small
\setlength{\tabcolsep}{6pt}
\renewcommand{\arraystretch}{1.15}
\begin{tabular}{lcccc}
\toprule
Alignment Reward & Math & BFCL & RAG \\
\midrule
Ordered matching      & 26.98 & 30.21 & 19.48 \\
Multiset matching     & 27.21 & 30.74 & 20.18 \\
Recall-only           & 27.05 & 30.49 & 19.96 \\
F1 (graded set)       & 27.43 & 31.04 & 20.61  \\
Set matching (\ours{}) & \textbf{27.95} & \textbf{31.52} & \textbf{21.08} \\
\bottomrule
\end{tabular}
\caption{Comparison of alternative flexible teacher-alignment rewards on Qwen2.5-7B. All variants are combined with \(R_c\) and \(R_v\) under the same GRPO setup. Math, BFCL, and RAG are overall task scores. Following the sweep protocol of \autoref{tab:ablation_table}, each configuration is independently trained once and scores are reported as means over three evaluation runs, whereas \autoref{tab:main_results} reports means over ten runs. The best score in each column is in bold.}
\label{tab:flexible_reward_comparison}
\end{table}

\paragraph{Why We Adopt Set Matching.}
Although the flexible variants exhibit similar training dynamics, their final task performance differs. As shown in Table~\ref{tab:flexible_reward_comparison}, set matching achieves the best overall performance across Math, BFCL, and RAG. This supports our choice of set matching as the default alignment reward in \ours{}.

The comparison also clarifies the trade-offs among alternatives. Ordered matching retains an order constraint, which can penalize valid trajectories that use the teacher's tools in a different executable order. Multiset matching relaxes ordering but requires the student to match the teacher's call counts, which can still penalize benign variation in tool use. Recall-only rewards coverage of the teacher's tools, but does not discourage additional unnecessary tools. F1 provides a graded overlap signal and performs closest to set matching, but its partial-credit structure can weaken the pressure to recover the full teacher tool set. By contrast, set matching focuses on recovering the teacher's core tool-selection behavior while remaining agnostic to order and call counts. This gives the student flexibility in trajectory construction while preserving a clear alignment target.

\paragraph{Implication.}
These results suggest that the contribution of \ours{} should be understood as a flexible teacher-alignment framework rather than as a method tied exclusively to one particular matching rule. The learning curves in Figure~\ref{fig:b3_flexible_variants} indicate that multiple flexible alignment rewards can address the tool-use trade-off between outcome-only and strict matching. Among these alternatives, Table~\ref{tab:flexible_reward_comparison} shows that set matching provides the strongest empirical trade-off in our setting. Other alignment functions may be appropriate in environments where partial overlap, call counts, or ordering information are more important.

\section{Details for Experimental Setup}
\subsection{Training Dataset Details \label{appen:training_data}}
\paragraph{Mathematical Reasoning as a Controlled Testbed.}
We use mathematical reasoning as the training domain because it provides a controlled testbed for studying reward design in tool-use distillation. Mathematical problems have well-defined answers, structured intermediate computations, and clear feedback from executable tools, allowing us to evaluate final correctness, tool validity, and tool-selection behavior with relatively low ambiguity. This helps isolate the effect of the reward structure from confounding factors such as noisy observations, underspecified goals, retrieval quality, corpus coverage, or environment-specific interaction failures.

We do not train separately on RAG or other domains because our goal is not to conduct an exhaustive cross-domain training study, but to test whether reward learning in a controlled tool-use setting transfers to unseen tool-use scenarios. Retrieval-based tasks introduce additional factors such as document availability, query formulation quality, and evidence retrieval errors, which can obscure whether failures arise from the reward design or from the external environment. We therefore use RAG and BFCL-v4 as out-of-domain evaluation benchmarks rather than additional training domains.

Mathematical reasoning is also sufficiently challenging for capacity-limited SLMs, as solving these problems often requires multi-step reasoning, appropriate tool invocation, and recovery from invalid or unnecessary tool calls. Thus, it provides a useful setting for examining the reward dilemma discussed in Section~\ref{sec:observation}.

\paragraph{Trajectory Construction.}
To train our models for effective tool use, we construct a high-quality dataset of reference trajectories, as described in Section~\ref{sec:experiment-setup}. To this end, we use the nvidia/AceReason-Math\footnote{\url{https://hf.co/datasets/nvidia/AceReason-Math}}~\citep{acereason} dataset as our source, leveraging its verified questions and ground-truth answers to ensure reliability.

The construction process involves providing these questions to our teacher model, Qwen3-235B, which generates solution trajectories using a calculator tool. We then filter these outputs, retaining only the successful trajectories in which the teacher's final answer matches the ground truth from the source dataset. This verification process yields a final training set of 1.27k single- and multi-turn trajectories that exemplify successful tool use.

\subsection{Versions of Models\label{appen:models}}
We employ Qwen3-235B-Thinking~\cite{qwen3technicalreport} as the teacher model and four student models from both Qwen3~\cite{qwen3technicalreport} and Qwen2.5~\cite{qwen2.5-math} families. The specific model versions and their Hugging Face identifiers are listed in Table~\ref{tab:model_versions}.

\begin{table}[h]
\centering
\footnotesize
\begin{tabular}{@{}llp{3.5cm}@{}}
\toprule
\textbf{Role} & \textbf{Model} & \textbf{Hugging Face Identifier} \\
\midrule
Teacher & Qwen3-235B-Thinking & \texttt{Qwen/Qwen3-235B-A22B- Thinking-2507-FP8} \\
\midrule
\multirow{4}{*}{Student} 
& Qwen2.5-1.5B-Instruct & \texttt{Qwen/Qwen2.5-1.5B -Instruct} \\
& Qwen2.5-7B-Instruct & \texttt{Qwen/Qwen2.5-7B -Instruct} \\
& Qwen3-1.7B & \texttt{Qwen/Qwen3-1.7B} \\
& Qwen3-8B & \texttt{Qwen/Qwen3-8B} \\
\bottomrule
\end{tabular}
\caption{Model versions used in experiments}
\label{tab:model_versions}
\end{table}

\subsection{Tool Execution via Remote Server\label{appen:math_tool}}
All tools described in Tables~\ref{tab:math_functions}, and~\ref{tab:rag_tool_function} are implemented as executable Python code. For tool execution, we use a FastAPI-based remote execution server, following the base architecture of the ReCall framework~\cite{chen2025learning}. Our implementation code is publicly available\footnote{\url{https://github.com/choics2623/MENTOR-RL}}.

When an agent invokes a tool, the system sends the tool's Python code to the remote server via HTTP API.
The server executes the code in a pre-configured environment with necessary libraries installed and returns the result.

\subsection{\label{appen:implementation}Implementation Details.}
The reinforcement learning framework is built on verl~\cite{hybridflow}. The agent is trained for two epochs on the combined training splits of our training dataset. We employ LoRA~\cite{hu2021lora} for supervised fine-tuning the student SLMs. For each model, we used the recommended sampling parameters from its official repository.
The `search' tool provided to the agent for the retrieval-based QA tasks is powered by a retrieval environment based on FlashRAG~\cite{FlashRAG}, using E5-base-v2~\cite{e5} as the retriever and the Dec. 2018 Wikipedia snapshot~\cite{karpukhin2020dense} as the knowledge base. For retrieval-based tasks, we retrieve the top-5 results for each query. 

Our training is conducted on $1\times8$ Nvidia H100 80G GPUs, with full parameter optimization and gradient checkpointing. We provide some important parameter settings in Tables~\ref{tab:implementation-details1} and ~\ref{tab:implementation-details2}.

\begin{table*}
\centering
\small
\begin{tabular}{@{}l >{\raggedright\arraybackslash}p{4cm} >{\raggedright\arraybackslash}p{3cm} >{\raggedright\arraybackslash}p{3.5cm}@{}}
\toprule
\textbf{Function} & \textbf{Description} & \textbf{Parameter Name} & \textbf{Parameter Description} \\
\midrule
\multirow{2}{*}{\texttt{add}} & \multirow{2}{4cm}{Add two numbers together} & 
\texttt{firstNumber} & The first number \\
& & \texttt{secondNumber} & The second number \\
\midrule
\multirow{2}{*}{\texttt{subtract}} & \multirow{2}{4cm}{Subtract one number from another} & 
\texttt{minuend} & The number to subtract from \\
& & \texttt{subtrahend} & The number to subtract \\
\midrule
\multirow{2}{*}{\texttt{multiply}} & \multirow{2}{4cm}{Multiply two numbers together} & 
\texttt{firstNumber} & The first number \\
& & \texttt{secondNumber} & The second number \\
\midrule
\multirow{2}{*}{\texttt{divide}} & \multirow{2}{4cm}{Divide one number by another} & 
\texttt{numerator} & The number to be divided \\
& & \texttt{denominator} & The number to divide by \\
\midrule
\texttt{sum\_numbers} & Calculate the sum of an array of numbers & 
\texttt{numbers} & Array of numbers to sum \\
\midrule
\texttt{floor} & Calculate the floor of a number & 
\texttt{number} & Number to find the floor of \\
\midrule
\texttt{ceil} & Calculate the ceil of a number & 
\texttt{number} & Number to find the ceil of \\
\midrule
\texttt{round\_number} & Round a number to the nearest integer & 
\texttt{number} & Number to round \\
\midrule
\multirow{2}{*}{\texttt{power}} & \multirow{2}{4cm}{Calculate base raised to the power of exponent} & 
\texttt{base} & The base number \\
& & \texttt{exponent} & The exponent \\
\midrule
\texttt{sqrt} & Calculate the square root of a number & 
\texttt{number} & Number to find the square root of \\
\midrule
\texttt{abs\_value} & Calculate the absolute value of a number & 
\texttt{number} & Number to find the absolute value of \\
\midrule
\multirow{2}{*}{\texttt{modulo}} & \multirow{2}{4cm}{Calculate the modulo of two numbers} & 
\texttt{dividend} & The dividend \\
& & \texttt{divisor} & The divisor \\
\bottomrule
\end{tabular}
\caption{Math Tool Functions\label{tab:math_functions}}
\end{table*}
\begin{table*}
\small
\centering
\begin{tabular}{@{}l >{\raggedright\arraybackslash}p{4.5cm} >{\raggedright\arraybackslash}p{3.5cm} >{\raggedright\arraybackslash}p{3.5cm}@{}}
\toprule
\textbf{Function} & \textbf{Description} & \textbf{Parameter Name (Type)} & \textbf{Parameter Description} \\
\midrule
\multirow{2}{*}{\texttt{wikipedia\_search}} & \multirow{2}{4.5cm}{Search Wikipedia for a given query.} & 
\texttt{query} (string) & Query to search for. \\
& & \texttt{top\_n} (integer) & Number of results to return. (Optional, default: 5) \\
\bottomrule
\end{tabular}
\caption{RAG Tool Function\label{tab:rag_tool_function}}
\end{table*}


\begin{table}[htbp]
\begin{center}
\small
\begin{tabular}{l|l}
\toprule
\textbf{Parameter} & \textbf{Value} \\
\midrule
Learning Rate & 1e-6 \\
Optimizer & AdamW \\
Epochs & 2 \\
Train Batch Size & 16 \\
Mini-batch Size & 8 \\
Max Sequence Length & 32768 \\
Max Response Length & 8192 \\
Number of Rollout & 10 \\
Tensor Model Parallel Size & 2 \\
Rollout Temperature (Qwen3) & 0.6 \\
Rollout Temperature (Qwen2.5) & 0.7 \\
GPU Utilization Ratio & 0.8 \\
KL Loss Coefficient & 0.001 \\
Clip Ratio & 0.2 \\
Reward Weights ($w_c, w_a, w_v$) & 0.7, 0.3, 0.1 \\
\bottomrule
\end{tabular}
\caption{Implementation details of \ours{}.\label{tab:implementation-details1}}
\end{center}
\vspace{-0.2in}
\end{table}

\begin{table}[htbp]
\begin{center}
\small
\begin{tabular}{l|l}
\toprule
\textbf{Parameter} & \textbf{Value} \\
\midrule
Learning Rate & 1e-7 \\
Optimizer & AdamW \\
Epochs & 1 \\
Train Batch Size & 4 \\
Gradient Accumulation Steps & 16 \\
Weight decay & 0.033 \\
Gradient Accum & 8 \\
\bottomrule
\end{tabular}
\caption{Implementation details of SFT.\label{tab:implementation-details2}}
\end{center}
\vspace{-0.2in}
\end{table}

\subsection{\label{appen:baseline_implementation}Implementation of the Baseline Models}
\paragraph{Outcome-Based RL.} To implement the Outcome-Based RL baseline, we adopt the reward structure of ReTool~\cite{feng2025retool}, which provides reinforcement signals exclusively upon successful final task completion. While the original ReTool framework utilizes Proximal Policy Optimization (PPO), we adapt this approach to the GRPO framework to ensure a fair algorithmic comparison. Under this setup, the reward configuration for the baseline is solely comprised of our correctness reward component ($R_c$), defined as:
\begin{equation}
\small
    R_{\text{c}} =
    \begin{cases}
    1 & \text{if } \hat{y}^{(s)} =  \hat{y}^{(t)} \\[6pt]
    0 & \text{otherwise}
    \end{cases}
\end{equation}
By removing all process-level feedback, this baseline isolates the student's ability to navigate the tool-use action space when guided only by the final outcome.

\paragraph{Strict RL.} To establish the Rigid RL baseline, we adapt the strict trajectory-matching reward mechanism introduced in ToolRL~\cite{qian2025toolrl}. While the original ToolRL framework enforces strict sequential matching against curated gold-standard ground-truth trajectories, such explicit human-annotated labels are absent in our distillation setup. Therefore, we adapt the mechanism to use the teacher-generated trajectory $\tau^{(t)}$ as the absolute reference in our GRPO optimization process. Unlike our flexible strategic alignment ($R_a$), which evaluates tool usage as an unordered set, the rigid reward ($R_{\text{strict format}}$) enforces a strict, position-wise sequential alignment. This requires the student's entire trajectory $\tau^{(s)}$ to exactly match the teacher's reference trajectory step-by-step, encompassing identical tool selection, argument values, and execution order:
\begin{equation}\small R_{\text{strict format}} = \begin{cases} 1 & \text{if } \tau^{(s)} = \tau^{(t)} \\[6pt]
0 & \text{otherwise} \end{cases}\end{equation}
Any deviation—such as altering the execution order, using a different tool, or generating mismatched arguments—results in a zero reward for that alignment component. This strict formulation intentionally bottlenecks exploratory autonomy, allowing us to accurately isolate the limitations of mechanical trajectory imitation.

\subsection{Benchmarks\label{appen:benchmark}}
\begin{table}[t]
\centering
\scriptsize
\setlength{\tabcolsep}{2pt} 
\renewcommand{\arraystretch}{0.9} 
\begin{tabular}{lllr}
\toprule
\textbf{Task Type} & \textbf{Dataset Name} &  \textbf{Description} & \textbf{Size} \\
\midrule
\multirow{6}{*}{\makecell{Math\\Reasoning}} 
    & Math-Forge-Hard            & College & 500 \\
    & Omni-MATH-512~\citep{gao2024omni}  & Olympiad & 512 \\
    & AIME24~\citep{AIME}         & Olympiad & 30 \\
    & AIME25                      & Olympiad & 30 \\
    & amc23~\citep{AMC2023_12AB}   & Olympiad & 40 \\
    & minervamath~\citep{minervamath}                 & College & 272 \\
\midrule
\multirow{1}{*}{\makecell{Tool-Calling}} 
    & BFCL v4~\citep{patil2025bfcl}        & Tool-call & 5088 \\
\midrule
\multirow{3}{*}{\makecell{Factual\\Reasoning}} 
    & HotPotQA~\citep{HotpotQA}   & 2-hop QA & 2000 \\
    & 2WikiMultiHopQA~\citep{2wiki} & 2-hop QA & 2000 \\
    & Bamboogle~\citep{press2023measuring} & 2-hop QA & 125 \\
\bottomrule
\end{tabular}
\caption{\label{tab:benchmarks} Benchmarks categorized by in-domain (\textit{Mathematical Reasoning}) and out-of-domain (\textit{Tool-Calling} and \textit{Factual Reasoning}) tasks and their test data size.}
\vspace{-0.15in} 
\end{table}

We evaluate our framework on a set of in-domain and out-of-domain tasks, detailed in \autoref{tab:benchmarks}, to measure both task-specific performance and generalization ability.
For in-domain tasks, our agent is trained and evaluated on a collection of mathematical reasoning benchmarks that provide a natural gradient of difficulty. This includes the widely-used MATH dataset~\cite{MATH}\footnote{\url{https://hf.co/datasets/prithivMLmods/Math-Forge-Hard}}, the tool-centric Omni-MATH-512~\cite{gao2024omni}\footnote{\url{https://hf.co/datasets/Heng1999/Omni-MATH-512}}, and several Olympiad-level datasets such as AIME24\footnote{\url{https://hf.co/datasets/math-ai/aime24}}, AIME25\footnote{\url{https://hf.co/datasets/math-ai/aime25}}, amc23\footnote{\url{https://hf.co/datasets/math-ai/amc23}}, and minervamath\footnote{\url{https://hf.co/datasets/math-ai/minervamath}}.
For out-of-domain tasks, we evaluate the trained agent on tasks requiring tools unseen during training to assess zero-shot generalization. To test \textit{retrieval-based QA}, we reframe multi-hop QA datasets (HotpotQA~\cite{HotpotQA}, 2WikiMultiHopQA~\cite{2wiki}, Bamboogle~\cite{press2023measuring}) as a tool-use problem where the agent is provided with a search(query) tool and must learn to call it effectively. To test broader capabilities, we also use the general \textit{Tool-Calling} benchmark BFCL v4~\cite{patil2025bfcl}, which involves a diverse set of novel tools.

\subsection{Teacher-Student Alignment Metric\label{appen:metric}}
To quantify alignment in tool usage patterns, we compare the distributions of tool calls across the 12 available tools, which are defined in \autoref{tab:math_functions}. Given the teacher's tool distribution $P$ and a student's distribution $Q$, we use an alignment score based on the Jensen-Shannon Divergence (JSD), defined as:
\begin{equation} \label{eq:alignment}
\small
\begin{gathered}
  \text{Alignment}(P, Q) = 1 - \text{JSD}(P, Q) \\
  \text{where } \text{JSD}(P, Q) = \sqrt{\frac{1}{2}\mathbb{D}_{KL}(P \| M) + \frac{1}{2}\mathbb{D}_{KL}(Q \| M)} \\
  \text{and } M = \frac{1}{2}(P + Q)
\end{gathered}
\end{equation}
The score is bounded between 0 and 1, where 1 signifies a perfect match.
$\mathbb{D}_{KL}$ denotes the Kullback-Leibler divergence.

Since our objective is tool-use distillation rather than independent policy discovery, teacher alignment serves as a diagnostic measure of capability transfer, not an absolute criterion for policy optimality. Importantly, we use task accuracy and exact match as the primary evaluation metrics, while the alignment score is used to assess whether performance gains are accompanied by successful transfer of tool-use patterns.

\paragraph{Interpretation of Alignment Score.}
The alignment score is intended as a diagnostic measure for analyzing tool-use transfer in the distillation setting, rather than as an absolute measure of policy optimality. Since our goal is to transfer tool-use capabilities from a stronger teacher model to smaller student models, similarity to the teacher's tool-usage distribution provides a useful behavioral signal for whether the student has acquired teacher-guided tool-selection patterns. However, high alignment alone does not imply that a policy is optimal; therefore, we report AS alongside downstream task performance, using accuracy and exact match as the primary evaluation metrics.

\section{Supplementary Results and Discussions}

\subsection{\label{appen:extensibility}Model Extensibility of Our Framework}
Because \ours{} defines alignment in terms of tool-use behavior rather than token-level distributions, its reward design is not tied to a specific tokenizer or model family. We therefore examine whether the same framework can be applied to student models outside the Qwen family. Specifically, we evaluate two LLaMA students using the same Qwen3-235B-Thinking teacher and report the results in \autoref{tab:cross-family}.

\paragraph{Cross-Family Setup.}
We retain the original Qwen3-235B-Thinking teacher and pair it with two LLaMA students that are close in scale to the students used in our main experiments: LLaMA-3.2-3B-Instruct and LLaMA-3.1-8B-Instruct. These LLaMA checkpoints are not reasoning-oriented tool-use models and show weaker initial TIR behavior in our sandbox environment. Therefore, directly comparing RL methods from the raw instruction-tuned checkpoints would primarily test whether the model can discover the executable tool-call interface, rather than whether a reward design improves tool-use distillation.

To control for this cold-start confound, we use a two-stage SFT-then-RL protocol for all RL-based methods. Specifically, we first apply a brief SFT warm-up on the same 1.27k filtered teacher trajectories to provide a minimal tool-use initialization, and then apply the method-specific RL objective for Outcome-based RL, Strict RL, and \ours{}. This warm-up is shared across all RL variants and is not specific to \ours{}. The Vanilla and SFT entries follow the same protocol as in the main experiments. For a controlled comparison, we keep the remaining hyperparameters, including rollout count, learning rate, KL coefficient, and reward weights, the same as in the Qwen experiments (Appendix~\ref{appen:implementation}).

\begin{table}[t]
\centering
\scriptsize
\setlength{\tabcolsep}{4pt}
\renewcommand{\arraystretch}{1.15}
\begin{tabular}{llccc}
\toprule
Model & Method & Math & BFCL & RAG \\
\midrule
\multirow{5}{*}{LLaMA-3.2-3B}  & Vanilla            & 2.85           & 22.18          & 1.92          \\
                               & SFT                & 5.62           & 23.94          & 2.74          \\
                               & Outcome-based RL   & 8.04           & 25.62          & 5.81          \\
                               & Strict RL          & 8.46           & 25.43          & 5.62          \\
                               & \ours{}      & \textbf{11.27} & \textbf{26.31} & \textbf{7.84} \\
\midrule
\multirow{5}{*}{LLaMA-3.1-8B}  & Vanilla            & 15.84          & 23.42          & 9.86          \\
                               & SFT                & 18.06          & 24.78          & 10.92         \\
                               & Outcome-based RL   & 20.42          & 28.71          & 13.74         \\
                               & Strict RL          & 20.78          & 28.92          & 13.92         \\
                               & \ours{}      & \textbf{22.71} & \textbf{30.46} & \textbf{15.62} \\
\bottomrule
\end{tabular}
\caption{Cross-family extensibility of \ours{}. Two LLaMA students are trained with the Qwen3-235B teacher using the two-stage SFT-then-RL pipeline described in the text. The best score in each column is in bold.}
\label{tab:cross-family}
\end{table}

\paragraph{Results.}
Table~\ref{tab:cross-family} reports the overall scores on the three benchmark families used in the main paper. For both LLaMA students, \ours{} achieves the best score across all benchmarks. The two RL baselines also improve over SFT in most cases, but remain below \ours{}, mirroring the general ordering observed in the Qwen experiments.

\paragraph{Observations.}
Three observations can be drawn from Table~\ref{tab:cross-family}. First, \ours{} consistently outperforms the Vanilla and SFT baselines across both LLaMA scales, suggesting that the proposed reward design can provide useful guidance even when the student model belongs to a different model family than the teacher. Second, \ours{} also outperforms both RL baselines on all three benchmark families, indicating that the gains are not simply due to applying RL after SFT warm-up, but are associated with the flexible, teacher-guided reward structure. Third, the absolute scores of the LLaMA students remain lower than those of the closest Qwen counterparts in Table~\ref{tab:main_results}. This is expected, as the LLaMA checkpoints used here are not reasoning-oriented models and show weaker initial TIR behavior in our sandbox environment. These results support our discussion in the Limitations section: initial tool-use ability can facilitate RL exploration, and a brief SFT warm-up provides only a partial initialization rather than a fully optimized TIR training procedure.

\paragraph{Discussion.}
The goal of this cross-family experiment is not to fully optimize LLaMA-based TIR agents, but to test whether the proposed reward design remains applicable when the student backbone differs from the teacher family. Under the shared SFT warm-up protocol, \ours{} consistently outperforms the SFT-only model and the two RL baselines across the three benchmark families. This indicates that the improvement is not merely due to exposing the LLaMA students to teacher trajectories, but comes from the subsequent flexible reward-based refinement.

At the same time, the absolute scores of the LLaMA students remain below those of the Qwen students in Table~\ref{tab:main_results}, suggesting that stronger initial TIR ability still matters. This result is consistent with our limitation discussion: for non-reasoning or weakly grounded backbones, cold-start training is a prerequisite for reliable RL exploration. We leave a more systematic study of cold-start data construction, tool-call format grounding, and model-family-specific TIR initialization for future work.

\subsection{\label{app:algorithm-extensibility}Algorithm Extensibility of Our Framework}
\ours{} is primarily a reward-design framework: its composite reward combines correctness, teacher alignment, and tool validation over completed rollouts (\autoref{eq:composite-reward}). While our main experiments use GRPO, the reward itself is not specific to GRPO's exact advantage formulation. We therefore examine whether the same reward design remains effective under related group-relative policy optimizers.

\paragraph{Alternative Algorithms.}
We compare GRPO with two recent variants from the same critic-free, group-relative optimization family. \textbf{DAPO}~\citep{yu2026dapo} modifies the optimization dynamics of GRPO through mechanisms such as decoupled clipping, dynamic sampling, token-level policy-gradient loss, and overlong-reward shaping. \textbf{GDPO}~\citep{liu2026gdpo} modifies reward normalization by normalizing each reward component before aggregation, making it particularly relevant for composite rewards such as \ours{}. These two variants allow us to test whether the benefit of \ours{} persists under different optimization dynamics and reward-normalization schemes.

\paragraph{Setup.}
We re-train Qwen2.5-7B under each optimizer with two reward configurations: the outcome-only reward \(R_c\) and the full \ours{} reward \(w_cR_c + w_aR_a + w_vR_v\), using the same weights as in the main experiments. All non-algorithm hyperparameters, including learning rate, batch size, KL coefficient, rollout count \(G=10\), training data, and evaluation benchmarks, are kept fixed as reported in Appendix~\ref{appen:implementation}. For DAPO and GDPO, we use their recommended algorithm-specific settings. Following the sweep protocol of Table~\ref{tab:ablation_table}, each configuration is independently trained once and evaluated over three runs; the GRPO rows therefore correspond to Settings (1) and (6) of that table.

\begin{table}[t]
\centering
\small
\setlength{\tabcolsep}{6pt}
\renewcommand{\arraystretch}{1.15}
\begin{tabular}{llccc}
\toprule
Algorithm & Reward design & Math & BFCL & RAG \\
\midrule
\multirow{2}{*}{GRPO} & Outcome-only  & 25.94          & 30.71          & 18.22          \\
                      & MENTOR (ours) & \textbf{27.95} & \textbf{31.52} & \textbf{21.08} \\
\midrule
\multirow{2}{*}{DAPO} & Outcome-only  & 25.97          & 31.02          & 18.28          \\
                      & MENTOR (ours) & \textbf{27.71} & \textbf{31.55} & \textbf{21.40} \\
\midrule
\multirow{2}{*}{GDPO} & Outcome-only  & 26.42          & 31.20          & 18.41          \\
                      & MENTOR (ours) & \textbf{28.19} & \textbf{31.79} & \textbf{21.62} \\
\bottomrule
\end{tabular}
\caption{Algorithm extensibility of \ours{}. The same composite reward is
applied under three policy-optimization algorithms on Qwen2.5-7B: GRPO
(critic-free, group-relative), DAPO (a large-scale GRPO variant), and GDPO
(a multi-reward-aware optimizer with per-component decoupled normalization).
Overall scores follow the metric conventions of Table~\ref{tab:main_results}.
Each configuration is independently trained once and evaluated over three runs,
whereas Table~\ref{tab:main_results} reports means over ten runs. The best score
within each algorithm group is in bold.}
\label{tab:algorithm-extensibility}
\end{table}

\begin{table*}[th]
\centering
\small
\setlength{\tabcolsep}{6pt}
\renewcommand{\arraystretch}{1.15}
\begin{tabular}{lccc}
\toprule
Stage & \ours{} & Standard RL & SFT \\
\midrule
Teacher trajectory generation (one-time) & $\sim$1\,h & --     & $\sim$1\,h \\
RL training                              & $\sim$6\,h & $\sim$6\,h & --        \\
SFT training                             & --         & --     & $\sim$2\,h    \\
\midrule
Reward computation overhead per step     & negligible & negligible & --   \\
\bottomrule
\end{tabular}
\caption{\label{tab:efficiency}Wall-clock cost of each training stage on $8 \times$~H100 (80\,GB).
Teacher trajectory generation is shared between \ours{} and SFT (both rely on
the same filtered teacher corpus) and is amortized across all subsequent
runs. Reward computation in \ours{} consists of set comparisons and binary
execution checks, which add negligible per-step overhead relative to the RL
rollout itself.}
\end{table*}

\paragraph{Results.}
Across all three optimizers, replacing the outcome-only reward with the full \ours{} reward consistently improves Math, BFCL, and RAG performance. The gains are similar across optimizers: \(+2.01\) / \(+0.81\) / \(+2.86\) for GRPO, \(+1.74\) / \(+0.53\) / \(+3.12\) for DAPO, and \(+1.77\) / \(+0.59\) / \(+3.21\) for GDPO on Math, BFCL, and RAG, respectively. This indicates that the benefit of \ours{} is not merely an artifact of the specific GRPO implementation used in the main experiments.

\paragraph{Discussion.}
These results provide preliminary evidence that \ours{} can be combined with multiple group-relative policy optimizers. The absolute differences among GRPO, DAPO, and GDPO are relatively small, and no single optimizer consistently dominates across all benchmarks. We therefore use GRPO in the main experiments because it offers a simple and competitive critic-free optimization backbone, allowing us to focus the comparison on reward design. Since these results are based on a single training run per configuration and remain within the group-relative optimizer family, a broader statistically controlled comparison across more policy-optimization algorithms is left for future work.

\paragraph{Scope of Algorithm Comparison.}
This comparison is not intended to cover all post-training algorithms. We focus on critic-free, group-relative online RL optimizers because they can consume the same rollout-level composite reward used in \ours{} with minimal changes to the training interface. Other algorithms, such as PPO or DPO, introduce additional design factors that would change the scope of the comparison. PPO typically requires a learned value function or critic and additional choices for value estimation and stabilization, while DPO is formulated as a preference-pair objective rather than direct optimization of scalar rollout rewards. Applying \ours{} under DPO would require a separate procedure for constructing chosen--rejected pairs from tool-use trajectories, making it a comparison of training paradigms rather than a controlled test of the reward design. We therefore leave broader comparisons across different post-training paradigms for future work and focus here on whether the proposed reward transfers across related group-relative optimizers.

\subsection{Analysis on Computational Cost}
RL-based distillation generally requires more training compute than SFT due to repeated rollout generation and policy optimization. The goal of this analysis is therefore not to claim that \ours{} is cheaper to train than SFT, but to clarify the additional cost introduced by our reward design relative to a standard outcome-only RL setup. In particular, \ours{} adds teacher-guided alignment rewards while keeping the underlying GRPO training procedure unchanged.

\paragraph{Cost Decomposition.}
Compared with vanilla outcome-only RL, \ours{} introduces two additional cost components. The first is a one-time preprocessing stage in which the teacher model generates reference trajectories used to construct the alignment signal \(R_a\). As described in Appendix~\ref{appen:method_detail}, the teacher \(\pi_{\text{teacher}}\) (Qwen3-235B-Thinking) is invoked on the AceReason-Math source questions, and the resulting trajectories are filtered against ground-truth answers, yielding a final corpus of \(1.27\)k successful tool-use trajectories.

The second component arises during training, where each student's rollout is scored against the corresponding reference trajectory. This step requires only lightweight operations, such as comparing the tools invoked in the student and teacher trajectories and checking whether tool calls execute successfully. Since these operations are computed outside the model forward and backward passes, their overhead is small relative to the cost of rollout generation and GRPO optimization.

\paragraph{Empirical Wall-Clock Comparison.}
Table~\ref{tab:efficiency} reports the wall-clock cost of each stage on the \(8 \times\) NVIDIA H100 (80\,GB) hardware used in our experiments. \ours{} and Standard RL share the same GRPO backbone with \(G = 10\) rollouts per question (Appendix~\ref{appen:method_detail}), so their RL training stages have comparable wall-clock costs. The main additional cost of \ours{} comes from the one-time teacher trajectory generation stage, which adds roughly \(1\) hour in our setup. This suggests that the proposed reward design does not substantially increase the cost of the RL optimization stage itself.

\paragraph{Inference-Time Implications.}
The practical motivation for distillation is to shift part of the cost from repeated large-model inference to a one-time training process for a smaller student model. \ours{} follows this motivation by improving the tool-use performance of capacity-limited SLMs (\(1.5\)B--\(8\)B parameters), as shown in Table~\ref{tab:main_results}. The actual cost benefit at deployment depends on the number of downstream queries, serving infrastructure, and the relative cost of teacher versus student inference. Nevertheless, the measured overhead indicates that \ours{} incurs only a modest additional training cost compared to standard RL, while enabling the use of smaller distilled models at inference time.

\section{Details of Instruction Prompts \label{appen:prompt}}
We utilize prompts for both reference trajectory generation and model evaluation. The specific prompt for reference trajectory generation is shown in \autoref{tab:prompt_qa_3_math}. For evaluation, we assess the robustness of \ours{} and the baseline models using prompts tailored to each of the three domains. All evaluation prompts for the Qwen2.5 and Qwen3 models are created by adapting the official chat templates provided with each model's tokenizer. The prompts are shown in Tables~\ref{tab:prompt_qa_2.5_math}, ~\ref{tab:prompt_qa_3_math}, ~\ref{tab:prompt_qa_2.5_rag}, and ~\ref{tab:prompt_qa_3_rag}. The instruction prompts used for evaluating the BFCL-v4 benchmark directly follow the format and content specified on the official benchmark website,\footnote{\url{https://gorilla.cs.berkeley.edu/blogs/15_bfcl_v4_web_search.html}} ensuring consistency with prior studies.\footnote{\url{https://github.com/ShishirPatil/gorilla/tree/main/berkeley-function-call-leaderboard}}

\begin{table*}[t] 
    \centering
    \small
    \begin{tcolorbox}[title={Prompt for Qwen2.5 + MATH}, width=0.9\textwidth]
        \textbf{SYSTEM:}
        \begin{lstlisting}[basicstyle=\small\ttfamily, breaklines=true]
system
You are Qwen, created by Alibaba Cloud. You are a helpful assistant.

# Tools

You may call one or more functions to assist with the user query.

You are provided with function signatures within <tools></tools> XML tags:
<tools>
{"type": "function", "function": {"name": "add", ...}}
{"type": "function", "function": {"name": "subtract", ...}}
{"type": "function", "function": {"name": "multiply", ...}}
...
{"type": "function", "function": {"name": "modulo", ...}}
</tools>

For each function call, return a json object with function name and arguments within <tool_call></tool_call> XML tags:

<tool_call>
{"name": <function-name>, "arguments": <args-json-object>}
</tool_call>

        \end{lstlisting}
        \vspace{1em} 
        \textbf{USER:}
        \begin{lstlisting}[basicstyle=\small\ttfamily, breaklines=true]
Question: Cities $A$ and $B$ are $45$ miles apart. Alicia lives in $A$ and Beth lives in $B$. Alicia bikes towards $B$ at 18 miles per hour. Leaving at the same time, Beth bikes toward $A$ at 12 miles per hour. How many miles from City $A$ will they be when they meet?

If you have got the answer, enclose it within \boxed{} with latex format.
        \end{lstlisting}
    \end{tcolorbox}
    \caption{Prompt for Qwen2.5 + MATH \label{tab:prompt_qa_2.5_math}}
\end{table*}
\begin{table*}[t] 
    \centering
    \small
    \begin{tcolorbox}[title={Prompt for Qwen3 + MATH}, width=0.9\textwidth]
        \textbf{SYSTEM:}
        \begin{lstlisting}[basicstyle=\small\ttfamily, breaklines=true]
system
# Tools

You may call one or more functions to assist with the user query.

You are provided with function signatures within <tools></tools> XML tags:
<tools>
{"type": "function", "function": {"name": "add", ...}}
{"type": "function", "function": {"name": "subtract", ...}}
{"type": "function", "function": {"name": "multiply", ...}}
...
{"type": "function", "function": {"name": "modulo", ...}}
</tools>

For each function call, return a json object with function name and arguments within <tool_call></tool_call> XML tags:
<tool_call>
{"name": <function-name>, "arguments": <args-json-object>}
</tool_call>
        \end{lstlisting}
        \vspace{1em} 
        \textbf{USER:}
        \begin{lstlisting}[basicstyle=\small\ttfamily, breaklines=true]
Question: Cities $A$ and $B$ are $45$ miles apart. Alicia lives in $A$ and Beth lives in $B$. Alicia bikes towards $B$ at 18 miles per hour. Leaving at the same time, Beth bikes toward $A$ at 12 miles per hour. How many miles from City $A$ will they be when they meet?

If you have got the answer, enclose it within \boxed{} with latex format.
        \end{lstlisting}
    \end{tcolorbox}
    \caption{Prompt for Qwen3 + MATH \label{tab:prompt_qa_3_math}}
\end{table*}
\begin{table*}[t] 
    \centering
    \small
    \begin{tcolorbox}[title={Prompt for Qwen2.5 + RAG}, width=0.9\textwidth]
        \textbf{SYSTEM:}
        \begin{lstlisting}[basicstyle=\small\ttfamily, breaklines=true]
system
You are Qwen, created by Alibaba Cloud. You are a helpful assistant.

# Tools

You may call one or more functions to assist with the user query.

You are provided with function signatures within <tools></tools> XML tags:
<tools>
{"type":"function", "function":{ 
  "name":"wikipedia_search",
  "description":"Search Wikipedia for a given query.",
  "parameters":{"type":"object", "properties":{
    "query":{"type":"string", "description":"Query to search for."},
    "top_n":{"type":"integer", "description":"Number of results to return. The default value is 5.", "default":5}}, 
   "required":["query"]}}}
</tools>

For each function call, return a json object with function name and arguments within <tool_call></tool_call> XML tags:

<tool_call>
{"name": <function-name>, "arguments": <args-json-object>}
</tool_call>
        \end{lstlisting}
        \vspace{1em} 
        \textbf{USER:}
        \begin{lstlisting}[basicstyle=\small\ttfamily, breaklines=true]
Question: What is the capital of France?

If you have got the answer, enclose it within \boxed{} with latex format.
        \end{lstlisting}
    \end{tcolorbox}
    \caption{Prompt for Qwen2.5 + RAG \label{tab:prompt_qa_2.5_rag}}
\end{table*}
\begin{table*}[t]  
    \centering
    \small
    \begin{tcolorbox}[title={Prompt for Qwen3 + RAG}, width=0.9\textwidth]
        \textbf{SYSTEM:}
        \begin{lstlisting}[basicstyle=\small\ttfamily, breaklines=true]
system
# Tools

You may call one or more functions to assist with the user query.

You are provided with function signatures within <tools></tools> XML tags:
<tools>
{"type":"function", "function":{ 
  "name":"wikipedia_search",
  "description":"Search Wikipedia for a given query.",
  "parameters":{"type":"object", "properties":{
    "query":{"type":"string", "description":"Query to search for."},
    "top_n":{"type":"integer", "description":"Number of results to return. The default value is 5.", "default":5}}, 
   "required":["query"]}}}
</tools>

For each function call, return a json object with function name and arguments within <tool_call></tool_call> XML tags:
<tool_call>
{"name": <function-name>, "arguments": <args-json-object>}
</tool_call>
        \end{lstlisting}
        \vspace{1em} 
        \textbf{USER:}
        \begin{lstlisting}[basicstyle=\small\ttfamily, breaklines=true]
Question: What is the capital of France?

If you have got the answer, enclose it within \boxed{} with latex format.
        \end{lstlisting}
    \end{tcolorbox}
    \caption{Prompt for Qwen3 + RAG \label{tab:prompt_qa_3_rag}}
\end{table*}

\end{document}